\documentclass[runningheads]{llncs}
\usepackage{graphicx}
\usepackage{amsmath,amssymb} 
\usepackage[utf8x]{inputenc}
\usepackage{pgfplots}
\pgfplotsset{compat=newest}
\usepackage{tikz}
\usepackage{color}
\usepackage{xcolor}
\usepackage{booktabs,siunitx} 
\usepackage{makecell}
\usepackage{enumitem}
\usepackage[caption=false]{subfig}
\usepackage[noadjust]{cite}
\usepackage[colorlinks=true,allcolors=green]{hyperref}
\usepackage{rotating}
\usepackage[misc]{ifsym}

\newcommand{\ie}{\mbox{\emph{i.e.\ }}}

\definecolor{road}                {RGB}{128, 64,128}
\definecolor{sidewalk}            {RGB}{244, 35,232}
\definecolor{building}            {RGB}{ 70, 70, 70}
\definecolor{wall}                {RGB}{102,102,156}
\definecolor{fence}               {RGB}{190,153,153}
\definecolor{pole}                {RGB}{153,153,153}
\definecolor{traffic light}       {RGB}{250,170, 30}
\definecolor{traffic sign}        {RGB}{220,220,  0}
\definecolor{vegetation}          {RGB}{107,142, 35}
\definecolor{terrain}             {RGB}{152,251,152}
\definecolor{sky}                 {RGB}{ 70,130,180}
\definecolor{person}              {RGB}{220, 20, 60}
\definecolor{rider}               {RGB}{255,  0,  0}
\definecolor{car}                 {RGB}{  0,  0,142}
\definecolor{truck}               {RGB}{  0,  0, 70}
\definecolor{bus}                 {RGB}{  0, 60,100}
\definecolor{train}               {RGB}{  0, 80,100}
\definecolor{motorcycle}          {RGB}{  0,  0,230}
\definecolor{bicycle}             {RGB}{119, 11, 32}
\definecolor{void}                {RGB}{  0,  0,  0}

\newcommand{\vecnorm}[1]{\left\|#1\right\|}

\newcommand{\best}[1]{\textbf{#1}}

\newcommand{\fnn}[1]{$\scriptstyle^{#1}$}

\begin{document}
\pagestyle{headings}
\mainmatter

\title{Model Adaptation with Synthetic and Real Data for Semantic Dense Foggy Scene Understanding}

\titlerunning{Model Adaptation with Synthetic and Real Foggy Data}

\authorrunning{C. Sakaridis \and D. Dai \and S. Hecker \and L. Van Gool}

\author{Christos Sakaridis\inst{1\text{(\Letter)}} \and Dengxin Dai\inst{1} \and Simon Hecker\inst{1} \and Luc Van Gool\inst{1,2}}

\institute{ETH Z\"urich, Z\"urich, Switzerland\\
	\email{ \{csakarid,dai,heckers,vangool\}@vision.ee.ethz.ch}
    \and
    KU Leuven, Leuven, Belgium
}

\maketitle

\begin{abstract}
This work addresses the problem of semantic scene understanding under dense fog. Although considerable progress has been made in semantic scene understanding, it is mainly related to clear-weather scenes. Extending recognition methods to adverse weather conditions such as fog is crucial for outdoor applications. In this paper, we propose a novel method, named Curriculum Model Adaptation (CMAda), which \emph{gradually} adapts a semantic segmentation model from light synthetic fog to dense real fog in multiple steps, using both synthetic and real foggy data. In addition, we present three other main stand-alone contributions: 1) a novel method to add synthetic fog to real, clear-weather scenes using semantic input; 2) a new fog density estimator; 3) the \emph{Foggy Zurich} dataset comprising $3808$ real foggy images, with pixel-level semantic annotations for $16$ images with dense fog. Our experiments show that 1) our fog simulation slightly outperforms a state-of-the-art competing simulation with respect to the task of semantic foggy scene understanding (SFSU); 2) CMAda improves the performance of state-of-the-art models for SFSU significantly by leveraging unlabeled real foggy data. The datasets and code will be made publicly available.

\end{abstract}

\section{Introduction}
\sloppy{Adverse weather conditions create visibility problems for both people and the sensors that power automated systems~\cite{vision:atmosphere,SFSU_synthetic,drive:surroundview:route:planner}. While sensors and the down-streaming vision algorithms are constantly getting better, their performance is mainly benchmarked with clear-weather images. Many outdoor applications, however, can hardly escape from bad weather. One typical example of adverse weather conditions is fog, which degrades the visibility of a scene significantly~\cite{contrast:weather:degraded,tan2008visibility}. The denser the fog is, the more severe this problem becomes.}

During the past years, the community has made a tremendous progress on image dehazing (defogging) to increase the visibility of foggy images ~\cite{bayesian:defogging,dark:channel,dehazing:mscale:depth}. The last few years have also witnessed a leap in object recognition. The semantic understanding of foggy scenes, however, has received little attention, despite its importance in outdoor applications. For example, an automated car still needs to detect other traffic agents and traffic control devices in the presence of fog. This work investigates the problem of semantic foggy scene understanding (SFSU).

The current ``standard'' policy for addressing semantic scene understanding is to train a neural network with many annotations of real images~\cite{imagenet:2015,Cityscapes}. Applying the same protocol to diverse weather conditions seems to be problematic, as the manual annotation part is hard to scale. The difficulty of data collection and annotation increases even more for adverse weather conditions. To overcome this problem, two streams of research have gained extensive attention: 1) transfer learning~\cite{DomainAdaptiveFasterRCNN} and 2) learning with synthetic data~\cite{Synthia:dataset,SFSU_synthetic}.

Our method falls into the middle ground, and aims to combine the strength of these two kinds of methods. In particular, our method is developed to learn from 1) a dataset with high-quality synthetic fog and corresponding human annotations, and 2) a dataset with a large number of images with real fog. The goal of our method is to improve the performance of SFSU without requiring extra human annotations.

\begin{figure}[t]
  \centering
  \includegraphics[width=0.95\linewidth]{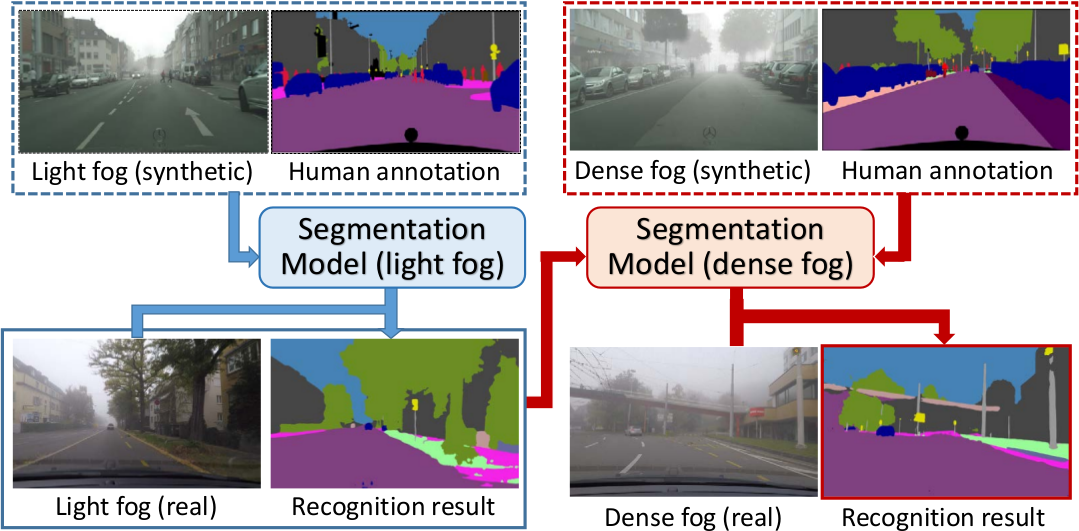}
  \caption{The illustrative pipeline of our approach for semantic scene understanding under dense fog}
  \label{fig:pipeline}
\end{figure}

To this aim, this work proposes a novel fog simulator to generate high-quality synthetic fog into real images that contain clear-weather outdoor scenes, and then leverage these partially synthetic foggy images for SFSU. The new fog simulator builds on the recent work in~\cite{SFSU_synthetic}, by introducing a semantic-aware filter to exploit the structures of object instances. We show that learning with our synthetic data improves the performance for SFSU. Furthermore, we present a novel method, dubbed Curriculum Model Adaptation (CMAda), which is able to \emph{gradually} adapt a segmentation model from light synthetic fog to dense real fog in multiple steps, by using both synthetic and real foggy data. CMAda improves upon direct adaptation significantly on two datasets with dense real fog.

The main contributions of the paper are: 1) a new automatic and scalable pipeline to generate high-quality synthetic fog, with which new datasets are generated; 2) a novel curriculum model adaptation method to learn from both synthetic and (unlabeled) real foggy images; 3) a new real foggy dataset with $3808$ images, including $16$ finely annotated images with dense fog. A visual overview of our approach is presented in Fig.~\ref{fig:pipeline}.

\section{Related Work}
Our work is relevant to image defogging (dehazing), foggy scene understanding, and domain adaptation.

\subsection{Image Defogging/Dehazing}
Fog fades the color of observed objects and reduces their contrast. Extensive research has been conducted on image defogging (dehazing) to increase the visibility of foggy scenes~\cite{contrast:weather:degraded,tan2008visibility,bayesian:defogging,fattal2008single,nonlocal:image:dehazing,fattal2014dehazing,dark:channel}. Certain works focus particularly on enhancing foggy road scenes~\cite{THC+12,exponential:contrast:restoration}. Recent approaches also rely on trainable architectures~\cite{TYW14}, which have evolved to end-to-end models~\cite{joint:transmission:estimation:dehazing,deep:transmission:network}. For a comprehensive overview of dehazing algorithms, we point the reader to~\cite{review:defogging:restoration,dehazing:survey:benchmarking}. Our work is complementary and focuses on semantic foggy scene understanding.

\subsection{Foggy Scene Understanding}
Typical examples in this line include road and lane detection~\cite{recent:progress:lane}, traffic light detection~\cite{traffic:light:survey:16}, car and pedestrian detection~\cite{kitti}, and a dense, pixel-level segmentation of road scenes into most of the relevant semantic classes~\cite{recognition:sfm:eccv08,Cityscapes}. While deep recognition networks have been developed~\cite{dilated:convolution,refinenet,pspnet,fast:rcnn,faster:rcnn} and large-scale datasets have been presented~\cite{kitti,Cityscapes}, that research mainly focused on clear weather. There is also a large body of work on fog detection~\cite{fog:detection:cv:09,fog:detection:vehicles:12,night:fog:detection,fast:fog:detection}. Classification of scenes into foggy and fog-free has been tackled as well~\cite{fog:nonfog:classification:13}. In addition, visibility estimation has been extensively studied for both daytime~\cite{visibility:road:fog:10,visibility:detection:fog:15,fog:detection:visibility:distance} and nighttime~\cite{night:visibility:analysis:15}, in the context of assisted and autonomous driving. The closest of these works to ours is~\cite{visibility:road:fog:10}, in which synthetic fog is generated and foggy images are segmented to \emph{free-space area} and \emph{vertical objects}. Our work differs in that our semantic understanding task is more complex and we tackle the problem from a different route by learning jointly from synthetic fog and real fog.

\subsection{Domain Adaptation}
Our work bears resemblance to transfer learning and model adaptation. Model adaptation across weather conditions to semantically segment simple road scenes is studied in~\cite{road:scene:2013}. More recently, domain adversarial based approaches were proposed to adapt semantic segmentation models both at pixel level and feature level from simulated to real environments~\cite{synthetic:semantic:segmentation,CyCADA}. Our work closes the domain gap by generating synthetic fog and by using the policy of \emph{gradual} adaptation. Combining our method and the aforementioned transfer learning methods is a promising direction. The concurrent work in~\cite{daytime:2:nighttime} on adaptation of semantic models from daytime to nighttime solely with real data is closely related to ours.

\section{Fog Simulation on Real Scenes Using Semantics}
\label{sec:simulation}

\begin{figure}[tb]
  \centering
  \includegraphics[width=\textwidth]{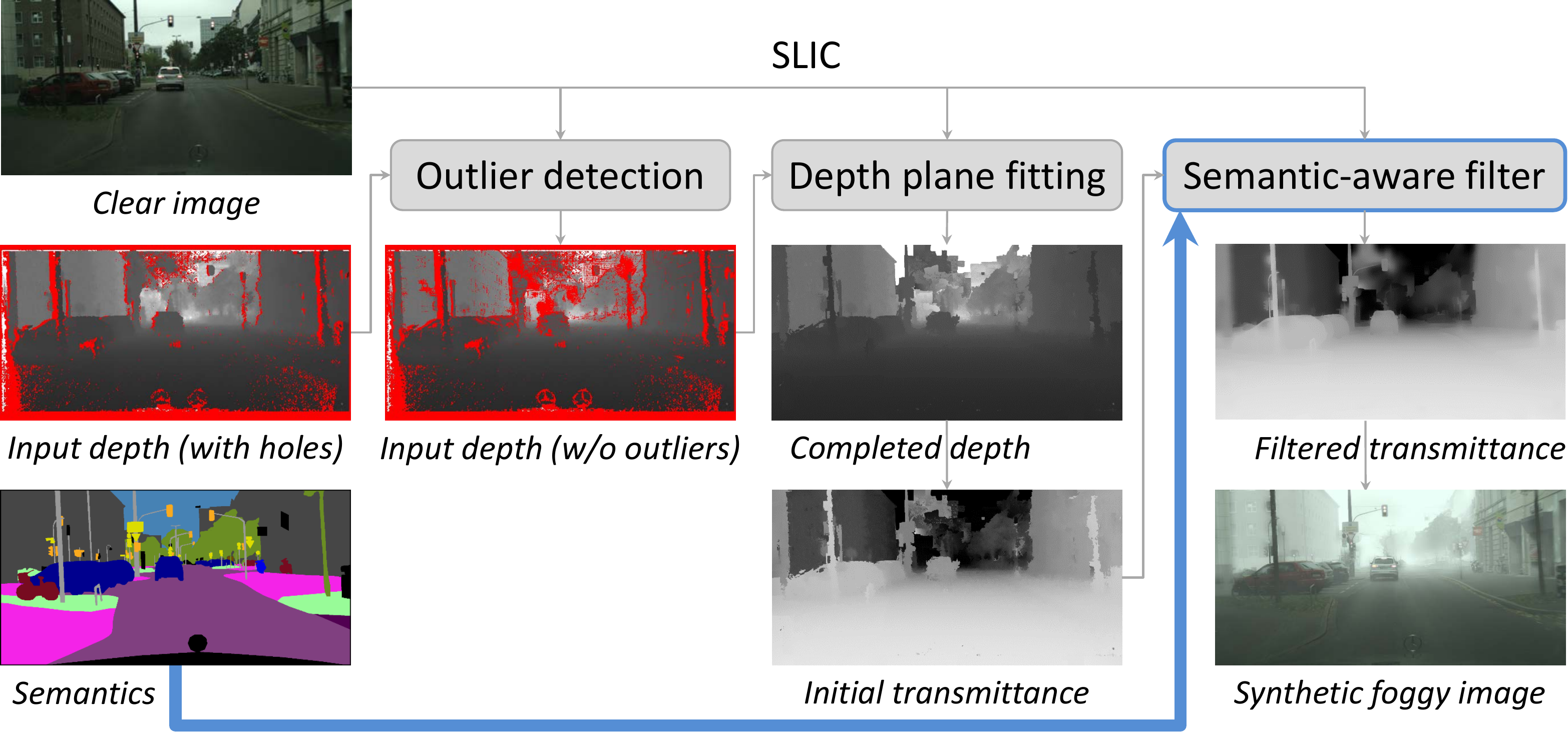}
  \caption{The pipeline of our fog simulation using semantics}
  \label{fig:simulation:pipeline}
\end{figure}

\subsection{Motivation}

We drive our motivation for fog simulation on real scenes using semantic input from the pipeline that was used in~\cite{SFSU_synthetic} to generate the Foggy Cityscapes dataset, which primarily focuses on depth denoising and completion. This pipeline is denoted in Fig.~\ref{fig:simulation:pipeline} with thin gray arrows and consists of three main steps: depth outlier detection, robust depth plane fitting at the level of SLIC superpixels~\cite{slic:superpixels} using RANSAC, and postprocessing of the completed depth map with guided image filtering~\cite{guided:filtering}. Our approach adopts the general configuration of this pipeline, but aims to improve its postprocessing step by leveraging the semantic annotation of the scene as additional reference for filtering, which is indicated in Fig.~\ref{fig:simulation:pipeline} with the thick blue arrow.

The guided filtering step in~\cite{SFSU_synthetic} uses the clear-weather color image as guidance to filter depth. However, as previous works on image filtering~\cite{mutual:structure:filtering} have shown, guided filtering and similar joint filtering methods such as cross-bilateral filtering~\cite{flash:enhance:crossbilateral:Eisemann,flash:enhance:crossbilateral:Petschnigg} transfer the structure that is present in the guidance/reference image to the output target image. Thus, any structure that is specific to the reference image but irrelevant for the target image is also transferred to the latter erroneously.

Whereas previous approaches such as mutual-structure filtering~\cite{mutual:structure:filtering} attempt to estimate the common structure between reference and target images, we identify this common structure with the structure that is present in the ground-truth \emph{semantic labeling} of the image. In other words, we assume that edges which are shared by the color image and the depth map generally coincide with \emph{semantic edges}, \ie{}locations in the image where the semantic classes of adjacent pixels are different. Under this assumption, the semantic labeling can be used directly as the reference image in a classical cross-bilateral filtering setting, since it contains exactly the mutual structure between the color image and the depth map. In practice, however, the boundaries drawn by humans in the semantic annotation are not pixel-accurate, and using the color image as additional reference helps to capture the precise shape of edges better. As a result, we formulate the postprocessing step of the completed depth map in our fog simulation as a \emph{dual-reference} cross-bilateral filter, with color and semantic reference.

\subsection{Dual-reference Cross-bilateral Filter Using Color and Semantics}
\label{sec:simulation:dual_bilateral}

Let us denote the RGB image of the clear-weather scene by $\mathbf{R}$ and its CIELAB counterpart by $\mathbf{J}$. We consider CIELAB, as it has been designed to increase perceptual uniformity and gives better results for bilateral filtering of color images~\cite{bilateral:grid}. The input image to be filtered in the postprocessing step of our pipeline constitutes a scalar-valued transmittance map $\hat{t}$. We provide more details on this transmittance map in Sec.~\ref{sec:simulation:rest}. Last, we are given a labeling function
\begin{equation} \label{eq:labeling}
h: \mathcal{P} \to \{1,\,\dots,\,C\}
\end{equation}
which maps pixels to semantic labels, where $\mathcal{P}$ is the discrete domain of pixel positions and $C$ is the total number of semantic classes in the scene. We define our dual-reference cross-bilateral filter with color and semantic reference as

\begin{equation} \label{eq:dual_bilateral}
t(\mathbf{p}) = \frac{\displaystyle\sum_{q \in \mathcal{N}(\mathbf{p})} G_{\sigma_s}(\vecnorm{\mathbf{q}-\mathbf{p}}) \left[\delta(h(\mathbf{q})-h(\mathbf{p})) + \mu G_{\sigma_c}(\vecnorm{\mathbf{J}(\mathbf{q})-\mathbf{J}(\mathbf{p})})\right] \hat{t}(\mathbf{q})}{\displaystyle\sum_{q \in \mathcal{N}(\mathbf{p})} G_{\sigma_s}(\vecnorm{\mathbf{q}-\mathbf{p}}) \left[\delta(h(\mathbf{q})-h(\mathbf{p})) + \mu G_{\sigma_c}(\vecnorm{\mathbf{J}(\mathbf{q})-\mathbf{J}(\mathbf{p})})\right]},
\end{equation}
where $\mathbf{p}$ and $\mathbf{q}$ denote pixel positions, $\mathcal{N}(\mathbf{p})$ is the neighborhood of $\mathbf{p}$, $\delta$ denotes the Kronecker delta, $G_{\sigma_s}$ is the spatial Gaussian kernel, $G_{\sigma_c}$ is the color-domain Gaussian kernel and $\mu$ is a positive constant. The novel dual reference is demonstrated in the second factor of the filter weights, which constitutes a sum of the terms $\delta(h(\mathbf{q})-h(\mathbf{p}))$ for semantic reference and $G_{\sigma_c}(\vecnorm{\mathbf{J}(\mathbf{q})-\mathbf{J}(\mathbf{p})})$ for color reference, weighted by $\mu$. The formulation of the semantic term implies that only pixels $\mathbf{q}$ with the same semantic label as the examined pixel $\mathbf{p}$ contribute to the output at $\mathbf{p}$ through this term, which prevents blurring of semantic edges. At the same time, the color term helps to better preserve true depth edges that do not coincide with any semantic boundary but are present in $\mathbf{J}$.

The formulation of~\eqref{eq:dual_bilateral} enables an efficient implementation of our filter based on the bilateral grid~\cite{bilateral:grid}. More specifically, we construct two separate bilateral grids that correspond to the semantic and color domains and operate separately on each grid to perform filtering, combining the results in the end. In this way, we handle a 3D bilateral grid for the semantic domain and a 5D grid for the color domain instead of a single joint 6D grid that would dramatically increase computation time~\cite{bilateral:grid}.

In our experiments, we set $\mu = 5$, $\sigma_s = 20$, and $\sigma_c = 10$.

\subsection{Remaining Steps}
\label{sec:simulation:rest}

Here we outline the rest parts of our fog simulation pipeline of Fig.~\ref{fig:simulation:pipeline}. For more details, we refer the reader to~\cite{SFSU_synthetic}, with which most parts of the pipeline are common. The standard optical model for fog that forms the basis of our fog simulation was introduced in~\cite{Koschmieder:optical:model} and is expressed as
\begin{equation} \label{eq:fog:model}
\mathbf{I}(\mathbf{x}) = \mathbf{R}(\mathbf{x})t(\mathbf{x}) + \mathbf{L}(1 - t(\mathbf{x})),
\end{equation}
where $\mathbf{I}(\mathbf{x})$ is the observed foggy image at pixel $\mathbf{x}$, $\mathbf{R}(\mathbf{x})$ is the clear scene radiance and $\mathbf{L}$ is the atmospheric light, which is assumed to be globally constant. The transmittance $t(\mathbf{x})$ determines the amount of scene radiance that reaches the camera. For homogeneous fog, transmittance depends on the distance $\ell(\mathbf{x})$ of the scene from the camera through
\begin{equation} \label{eq:transmittance}
t(\mathbf{x}) = \exp\left(-\beta\ell(\mathbf{x})\right).
\end{equation}
The attenuation coefficient $\beta$ controls the density of the fog: larger values of $\beta$ mean denser fog. Fog decreases the meteorological optical range (MOR), also known as visibility, to less than 1 km by definition~\cite{Federal:meteorological:handbook}. For homogeneous fog $\text{MOR}=2.996/\beta$, which implies
\begin{equation} \label{eq:beta:bound:fog}
\beta \geq 2.996\times{}10^{-3}{\text{ m}}^{-1},
\end{equation}
where the lower bound corresponds to the lightest fog configuration. In our fog simulation, the value that is used for $\beta$ always obeys \eqref{eq:beta:bound:fog}.

The required inputs for fog simulation with \eqref{eq:fog:model} are the image $\mathbf{R}$ of the original clear scene, atmospheric light $\mathbf{L}$ and a complete transmittance map $t$. We use the same approach for atmospheric light estimation as that in~\cite{SFSU_synthetic}. Moreover, we adopt the stereoscopic inpainting method of~\cite{SFSU_synthetic} for depth denoising and completion to obtain an initial complete transmittance map $\hat{t}$ from a noisy and incomplete input disparity map $D$, using the recommended parameters. We filter $\hat{t}$ with our dual-reference cross-bilateral filter~\eqref{eq:dual_bilateral} to compute the final transmittance map $t$, which is used in~\eqref{eq:fog:model} to synthesize the foggy image $\mathbf{I}$.

\begin{figure*}[tb]
    \centering
    \subfloat{\includegraphics[width=0.31\textwidth]{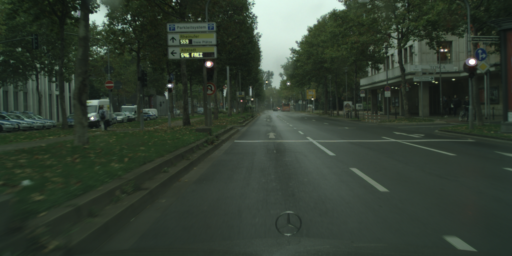}}
    \hfil
    \subfloat{\includegraphics[width=0.31\textwidth]{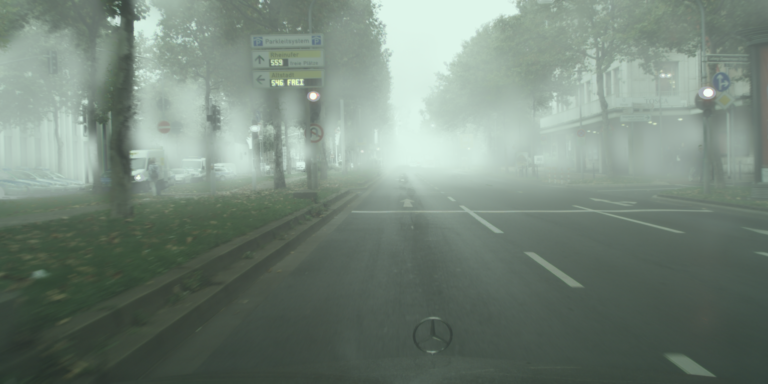}}
    \hfil
    \subfloat{\includegraphics[width=0.31\textwidth]{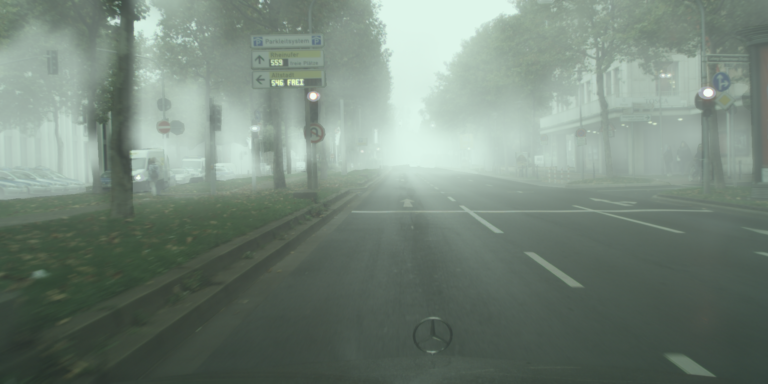}}
    \\
    \subfloat{\includegraphics[width=0.31\textwidth]{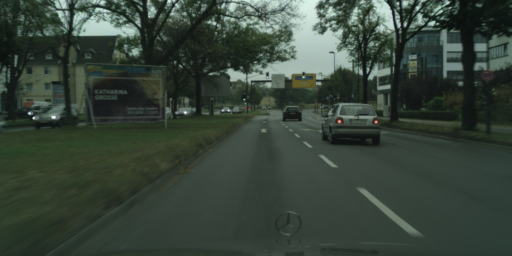}}
    \hfil
    \subfloat{\includegraphics[width=0.31\textwidth]{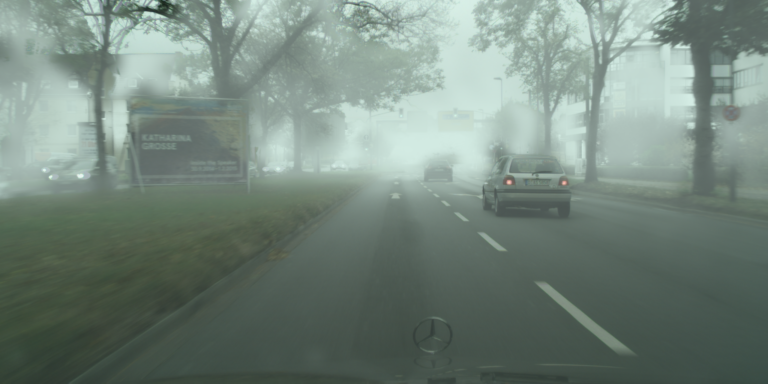}}
    \hfil
    \subfloat{\includegraphics[width=0.31\textwidth]{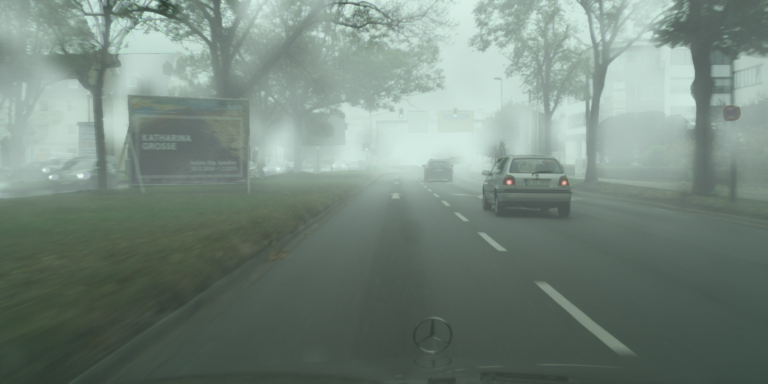}}
    \\
    \addtocounter{subfigure}{-6}
    \subfloat[Cityscapes]{\includegraphics[width=0.31\textwidth]{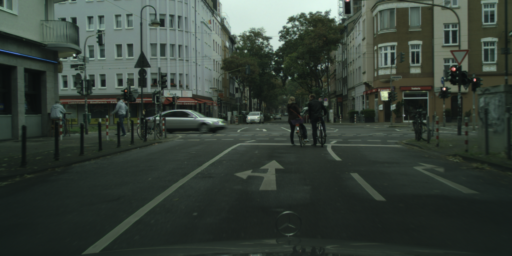}\label{fig:fog:simulation:input}}
    \hfil
    \subfloat[Foggy Cityscapes]{\includegraphics[width=0.31\textwidth]{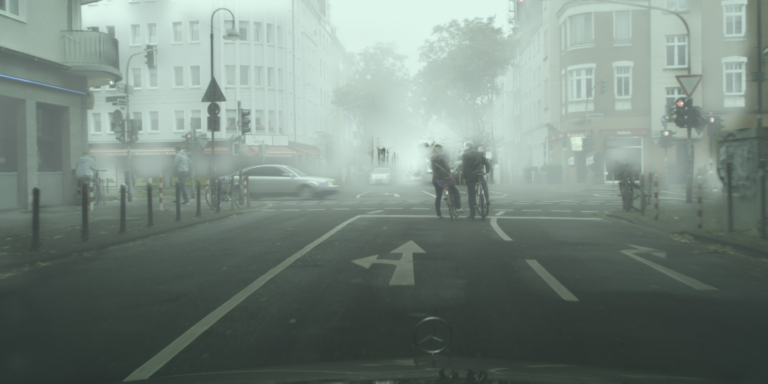}\label{fig:fog:simulation:sfsu_synthetic}}
    \hfil
    \subfloat[Our foggy image]{\includegraphics[width=0.31\textwidth]{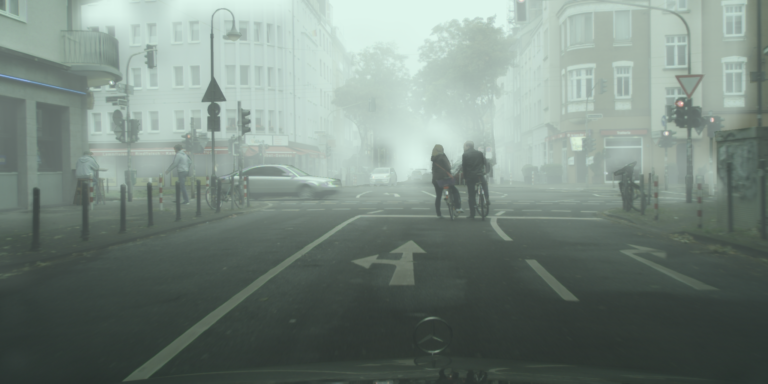}\label{fig:fog:simulation:ours}}
    \caption{Comparison of our synthetic foggy images against Foggy Cityscapes~\cite{SFSU_synthetic}. This figure is better seen on a screen and zoomed in}
    \label{fig:fog:simulation}
\end{figure*}

Results of the presented pipeline for fog simulation on example images from Cityscapes~\cite{Cityscapes} are provided in Fig.~\ref{fig:fog:simulation} for $\beta = 0.02$, which corresponds to visibility of ca.\ $150\text{m}$. We specifically leverage the instance-level semantic annotations that are provided in Cityscapes and set the labeling $h$ of~\eqref{eq:labeling} to a different value for each distinct instance of the same semantic class in order to distinguish adjacent instances. We compare our synthetic foggy images against the respective images of Foggy Cityscapes that were generated with the approach of~\cite{SFSU_synthetic}. Our synthetic foggy images generally preserve the edges between adjacent objects with large discrepancy in depth better than the images in Foggy Cityscapes, because our approach utilizes semantic boundaries, which usually encompass these edges. The incorrect structure transfer of color textures to the transmittance map, which deteriorates the quality of Foggy Cityscapes, is also reduced with our method.

\section{Semantic Segmentation of Scenes with Dense Fog}

In this section, we first present a standard supervised learning approach for semantic segmentation under dense fog using our synthetic foggy data with the novel fog simulation of Sec.~\ref{sec:simulation}, and then elaborate on our novel curriculum model adaptation approach using both synthetic and real foggy data.

\subsection{Learning with Synthetic Fog}

Generating synthetic fog from real clear-weather scenes grants the potential of inheriting the existing human annotations of these scenes, such as those from the Cityscapes dataset~\cite{Cityscapes}. This is a significant asset that enables training of standard segmentation models. Therefore, an effective way of evaluating the merit of a fog simulator is to adapt a segmentation model originally trained on clear weather to the synthesized foggy images and then evaluate the adapted model against the original one on real foggy images. The goal is to verify that the standard learning methods for semantic segmentation can benefit from our simulated fog in the challenging scenario of real fog. This evaluation policy has been proposed in~\cite{SFSU_synthetic}. We adopt this policy and fine-tune the RefineNet model~\cite{refinenet} on synthetic foggy images generated with our simulation. The performance of our adapted models on dense real fog is compared to that of the original clear-weather model as well as the models that are adapted on Foggy Cityscapes~\cite{SFSU_synthetic}, providing an objective comparison of our simulation method against~\cite{SFSU_synthetic}.

\subsection{Curriculum Model Adaptation with Synthetic and Real Fog}

While adapting a standard segmentation model to our synthetic fog improves its performance as shown in Sec.~\ref{sec:experiments:synthetic}, the paradigm still suffers from the domain discrepancy between synthetic and real foggy images. This discrepancy becomes more accentuated for denser fog. We present a method which can learn from our synthetic fog plus unlabeled real foggy data.

The method, which we term Curriculum Model Adaptation (CMAda), uses two versions of synthetic fog---one with light fog and another with dense fog---and a large dataset of unlabeled real foggy scenes with variable, unknown fog density, and works as follows:
\begin{enumerate}
\item generate a synthetic foggy dataset with multiple versions of varying fog density; \label{itemone}
\item train a model for fog density estimation on the dataset of step~\ref{itemone}; \label{itemtwo}
\item rank the images in the real foggy dataset with the model of step~\ref{itemtwo} according to fog density; \label{itemthree}
\item generate a dataset with light synthetic fog, and train a segmentation model on it; \label{itemfour}
\item apply the segmentation model from step~\ref{itemfour} to the light-fog images of the real dataset (ranked lower in step~\ref{itemthree}) to obtain ``noisy'' semantic labels; \label{itemfive}
\item generate a dataset with dense synthetic fog; \label{itemsix}
\item adapt the segmentation model from step~\ref{itemfour} to the union of the dense synthetic foggy dataset from step~\ref{itemsix} and the light real foggy one from step~\ref{itemfive}. \label{itemseven}
\end{enumerate}

CMAda adapts segmentation models from light synthetic fog to dense real fog and is inspired by curriculum learning~\cite{curriculum:learning}, in the sense that we first solve easier tasks with our synthetic data, \ie{}fog density estimation and semantic scene understanding under light fog, and then acquire new knowledge from the already ``solved'' tasks in order to better tackle the harder task, \ie{}scene understanding under dense real fog. CMAda also exploits the direct control of fog density for synthetic foggy images. Fig.~\ref{fig:pipeline} provides an overview of our method. Below we present details on our fog density estimation, \ie{}step~\ref{itemtwo}, and the training of the model, \ie{}step~\ref{itemseven}.

\subsubsection{Fog Density Estimation.}
\label{sec:fog:density:estimation}
Fog density is usually determined by the visibility of the foggy scene. An accurate estimate of fog density can benefit many applications, such as image defogging~\cite{fog:density:15}. Since annotating images in a fine-grained manner regarding fog density is very challenging, previous methods are trained on a few hundreds of images divided into only two classes: foggy and fog-free~\cite{fog:density:15}. The performance of the system, however, is affected by the small amount of training data and the coarse class granularity.

In this paper, we leverage our fog simulation applied to Cityscapes~\cite{Cityscapes} for fog density estimation. Since simulated fog density is directly controlled through $\beta$, we generate several versions of Foggy Cityscapes with varying $\beta \in \{0,\,0.005,\,0.01,\,0.02\}$ and train AlexNet~\cite{alexnet} to regress the value of $\beta$ for each image, lifting the need to handcraft features relevant to fog as \cite{fog:density:15} did. The predicted fog density using our method correlates well with human judgments of fog density taken in a subjective study on a large foggy image database on Amazon Mechanical Turk (cf.\ Sec.~\ref{sec:exp:fog:density} for results). The fog density estimator is used to rank our new \emph{Foggy Zurich} dataset, to select light foggy images for usage in CMAda, and to select dense foggy images for manual annotation.

\subsubsection{Curriculum Model Adaptation.}
\label{sec:CMAda}
We formulate CMAda for semantic segmentation as follows. Let us denote a clear-weather image by $\mathbf{x}$, the corresponding image under light synthetic fog by $\mathbf{x}^\prime$, the corresponding image under dense synthetic fog by $\mathbf{x}^{\prime \prime}$, and the corresponding human annotation by $\mathbf{y}$. Then, the training data consist of labeled data with light synthetic fog $\mathcal{D}^\prime_l =\{(\mathbf{x}^\prime_i, \mathbf{y}_i)\}_{i=1}^{l}$, labeled data with dense synthetic fog $\mathcal{D}^{\prime \prime}_l =\{(\mathbf{x}^{\prime \prime}_i, \mathbf{y}_i)\}_{i=1}^{l}$  and unlabeled images with light real fog $\mathcal{\bar{D}}^\prime_u =\{\mathbf{\bar{x}}^\prime_j\}_{j=l+1}^{l+u}$, where $\mathbf{y}_i^{m,n} \in\{1, ..., C\}$  is the label of pixel $(m,n)$, and $C$ is the total number of classes. $l$ is the number of labeled training images with synthetic fog, and $u$ is the number of unlabeled images with light real fog. The aim is to learn a mapping function  $\phi^{\prime \prime}: \mathcal{X}^{\prime \prime} \mapsto \mathcal{Y}$ from $\mathcal{D}^\prime_l$, $\mathcal{D}^{\prime \prime}_l$ and $\mathcal{\bar{D}}^\prime_u$, and evaluate it on images with dense real fog $\mathcal{\bar{D}}^{\prime \prime} = \{\mathbf{\bar{x}}^{\prime \prime}_1, \dots, \mathbf{\bar{x}}^{\prime \prime}_k\}$, where $k$ is the number of images with dense real fog.

Since $\mathcal{\bar{D}}^\prime_u$ does not have human annotations, we generate the supervisory labels as previously described in step~\ref{itemfive}. In particular, we first learn a mapping function $\phi^\prime: \mathcal{X^\prime} \mapsto \mathcal{Y}$ with $\mathcal{D}^{\prime}_l$ and then obtain the labels $\bar{\mathbf{y}}^\prime_j=\phi^\prime(\mathbf{\bar{x}}^\prime_j)$ for $\mathbf{\bar{x}}^\prime_j$, $\forall j \in \{l+1, \dots, l+u\}$. $\mathcal{\bar{D}}^\prime_u$ is then upgraded to $\mathcal{\bar{D}}^\prime_u=\{(\mathbf{\bar{x}}^\prime_j, \bar{\mathbf{y}}^\prime_j)\}_{j=l+1}^{l+u}$. The proposed scheme for training semantic segmentation models for dense foggy image $\bar{\mathbf{x}}^{\prime \prime}$ is to learn a mapping function  $\phi^{\prime \prime}$ so that human annotations for dense synthetic fog and the generated labels for light real fog are both taken into account:
\begin{equation}
\min_{\phi^{\prime \prime}} \frac{1}{l}\sum_{i=1}^l L(\phi^{\prime \prime}(\mathbf{x}^{\prime \prime}_i), \mathbf{y}_i) + \lambda \frac{1}{u}\sum_{j=l+1}^{l+u} L(\phi^{\prime \prime}(\mathbf{\bar{x}}^\prime_j), \bar{\mathbf{y}}^\prime_j),
\label{eq:ssl}
\end{equation}
where $L(.,.)$ is the cross entropy loss function and $\lambda=\frac{u}{l}\times w$ is a hyper-parameter balancing the weights of the two data sources, with $w$ serving as the relative weight of each real weakly labeled image compared to each synthetic labeled one. We empirically set $w=1/3$ in our experiment, but an optimal value can be obtained via cross-validation if needed. The optimization of \eqref{eq:ssl} is implemented by mixing images from $\mathcal{D}^{\prime \prime}_l$ and $\bar{\mathcal{D}}^\prime_u$  in a proportion of $1:w$ and feeding the stream of hybrid data to a CNN for standard supervised training.

This learning approach bears resemblance to model distillation~\cite{hinton2015distilling,supervision:transfer} or imitation~\cite{model:compression,dai:metric:imitation}. The underpinnings of our proposed approach are the following: 1) in light fog objects are easier to recognize than in dense fog, hence models trained on synthetic data are more generalizable to real data in case both data sources contain light rather than dense fog; 2) dense synthetic fog and light real fog reflect different and complementary characteristics of the target domain of dense real fog. On the one hand, dense synthetic fog features a similar overall visibility obstruction to dense real fog, but includes artifacts. On the other hand, light real fog captures the true nonuniform and spatially varying structure of fog, but at a different density than dense fog.

\section{The Foggy Zurich Dataset}

\subsection{Data Collection}

\emph{Foggy Zurich} was collected during multiple rides with a car inside the city of Zurich and its suburbs using a GoPro~Hero~5 camera. We recorded four large video sequences, and extracted video frames corresponding to those parts of the sequences where fog is (almost) ubiquitous in the scene at a rate of one frame per second. The extracted images are manually cleaned by removing the duplicates (if any), resulting in $3808$ foggy images in total. The resolution of the frames is $1920 \times 1080$ pixels. We mounted the camera inside the front windshield, since we found that mounting it outside the vehicle resulted in significant deterioration in image quality due to blurring artifacts caused by dew.

\subsection{Annotation of Images with Dense Fog}
\label{sec:dataset:annotations}

\begin{figure*}[tb]
    \centering
    \begin{tikzpicture}
    \tikzstyle{every node}=[font=\footnotesize]
    \begin{axis}[
      ybar,
      ymode=log,
      width=\textwidth,
      height=6cm,
      xmin=0,
      xmax=26,
      ymin=1,
      ymax=5e8,
      axis y discontinuity=crunch,
      ymajorgrids=true,
      ylabel={number of pixels},
      ytick={1,1e2,1e4,1e6,1e8},
      yticklabels={0,$10^2$,$10^4$,$10^6$,$10^8$,},
      xtick={1.5,5,8.5,13.5,18,21,24.5},
      minor xtick={3,7,10,17,19,23},
      xticklabels = {
        flat,
        construction,
        nature,
        vehicle,
        sky,
        object,
        human,
      },
      major x tick style = {opacity=0},
      minor x tick num = 1,
      xtick pos=left,
      every node near coord/.append style={
      anchor=west,
      rotate=90,
      font=\scriptsize,
      }
    ]

    \addplot[bar shift=0pt,draw=road,          fill opacity=0.9,fill=road!80!white           , nodes near coords=road                 ] plot coordinates{ ( 1,     4421067  ) };
    \addplot[bar shift=0pt,draw=sidewalk,      fill opacity=0.8,fill=sidewalk!80!white       , nodes near coords=sidewalk             ] plot coordinates{ ( 2,     662095   ) };

    \addplot[bar shift=0pt,draw=building,      fill opacity=0.8,fill=building!80!white       , nodes near coords=build.               ] plot coordinates{ ( 4,     2981421  ) };
    \addplot[bar shift=0pt,draw=fence,         fill opacity=0.8,fill=fence!80!white          , nodes near coords=fence                ] plot coordinates{ ( 5,     758634   ) };
    \addplot[bar shift=0pt,draw=wall,          fill opacity=0.8,fill=wall!80!white           , nodes near coords=wall                 ] plot coordinates{ ( 6,     1721511  ) };

    \addplot[bar shift=0pt,draw=vegetation,    fill opacity=0.8,fill=vegetation!80!white     , nodes near coords=veget.               ] plot coordinates{ ( 8,    2447052   ) };
    \addplot[bar shift=0pt,draw=terrain,       fill opacity=0.8,fill=terrain!80!white        , nodes near coords=terrain              ] plot coordinates{ ( 9,    140340    ) };

    \addplot[bar shift=0pt,draw=car,           fill opacity=0.8,fill=car!80!white            , nodes near coords=car\fnn{1}           ] plot coordinates{ ( 11,    193327   ) };
    \addplot[bar shift=0pt,draw=truck,         fill opacity=0.8,fill=truck!80!white          , nodes near coords=truck\fnn{1}         ] plot coordinates{ ( 12,    2285     ) };
    \addplot[bar shift=0pt,draw=train,         fill opacity=0.8,fill=train!80!white          , nodes near coords=train\fnn{1}         ] plot coordinates{ ( 13,    1        ) };
    \addplot[bar shift=0pt,draw=bus,           fill opacity=0.8,fill=bus!80!white            , nodes near coords=bus\fnn{1}           ] plot coordinates{ ( 14,    7475     ) };
    \addplot[bar shift=0pt,draw=bicycle,       fill opacity=0.8,fill=bicycle!80!white        , nodes near coords=bicycle\fnn{1}       ] plot coordinates{ ( 15,    383      ) };
    \addplot[bar shift=0pt,draw=motorcycle,    fill opacity=0.8,fill=motorcycle!80!white     , nodes near coords=motorcycle\fnn{1}    ] plot coordinates{ ( 16,    1        ) };

    \addplot[bar shift=0pt,draw=sky,           fill opacity=0.8,fill=sky!80!white            , nodes near coords=sky                  ] plot coordinates{ ( 18,    12313281 ) };

    \addplot[bar shift=0pt,draw=pole,          fill opacity=0.8,fill=pole!80!white           , nodes near coords=pole                 ] plot coordinates{ ( 20,    385192   ) };
    \addplot[bar shift=0pt,draw=traffic sign,  fill opacity=0.8,fill=traffic sign!80!white   , nodes near coords=traffic sign         ] plot coordinates{ ( 21,    239184   ) };
    \addplot[bar shift=0pt,draw=traffic light, fill opacity=0.8,fill=traffic light!80!white  , nodes near coords=traffic light        ] plot coordinates{ ( 22,    32330    ) };

    \addplot[bar shift=0pt,draw=person,        fill opacity=0.8,fill=person!80!white         , nodes near coords=person\fnn{1}        ] plot coordinates{ ( 24,    1139     ) };
    \addplot[bar shift=0pt,draw=rider,         fill opacity=0.8,fill=rider!80!white          , nodes near coords=rider\fnn{1}         ] plot coordinates{ ( 25,    1        ) };

    \node at (axis cs:16.5,7.5e8) [draw=none,anchor=north west,font=\scriptsize] {\fnn{1} instances are distinguished};

    \end{axis}
    \end{tikzpicture}
    \caption{Number of annotated pixels per class for \emph{Foggy Zurich-test}}
    \label{fig:dataset:stats:segmentation}
\end{figure*}
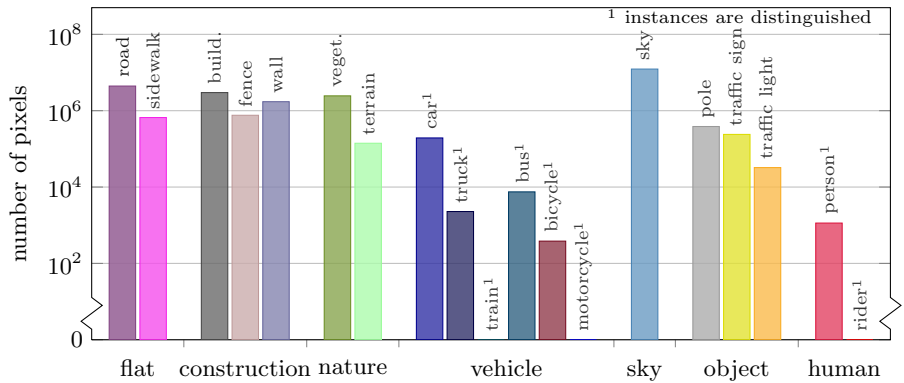

We use our fog density estimator presented in Sec.~\ref{sec:fog:density:estimation} to rank all images in \emph{Foggy Zurich} according to fog density.  Based on the ordering, we manually select 16 images with \emph{dense} fog and diverse visual scenes, and construct the test set of \emph{Foggy Zurich} therefrom, which we term \emph{Foggy Zurich-test}. We annotate these images with fine pixel-level semantic annotations using the 19 evaluation classes of the Cityscapes dataset~\cite{Cityscapes}. In addition, we assign the \emph{void} label to pixels which do not belong to any of the above 19 classes, or the class of which is uncertain due to the presence of fog. Every such pixel is ignored for semantic segmentation evaluation. Comprehensive statistics for the semantic annotations of \emph{Foggy Zurich-test} are presented in Fig.~\ref{fig:dataset:stats:segmentation}. We also distinguish the semantic classes that occur frequently in \emph{Foggy Zurich-test}. These ``frequent'' classes are: \emph{road}, \emph{sidewalk}, \emph{building}, \emph{wall}, \emph{fence}, \emph{pole}, \emph{traffic light}, \emph{traffic sign}, \emph{vegetation}, \emph{sky}, and \emph{car}. When performing evaluation on \emph{Foggy Zurich-test}, we occasionally report the average score over this set of frequent classes, which feature plenty of examples, as a second metric to support the corresponding results.

Despite the fact that there exists a number of prominent large-scale datasets for semantic road scene understanding, such as KITTI~\cite{kitti}, Cityscapes~\cite{Cityscapes} and Mapillary Vistas~\cite{Mapillary}, most of these datasets contain few or even no foggy scenes, which can be attributed partly to the rarity of the condition of fog and the difficulty of annotating foggy images. To the best of our knowledge, the only previous dataset for semantic foggy scene understanding whose scale exceeds that of \emph{Foggy Zurich-test} is Foggy Driving~\cite{SFSU_synthetic}, with 101 annotated images. However, we found that most images in Foggy Driving contain relatively light fog and most images with dense fog are annotated \emph{coarsely}. Compared to Foggy Driving, \emph{Foggy Zurich} comprises a much greater number of high-resolution foggy images. Its larger, unlabeled part is highly relevant for unsupervised or semi-supervised approaches such as the one we have presented in Sec.~\ref{sec:CMAda}, while the smaller, labeled \emph{Foggy Zurich-test} set features \emph{fine} semantic annotations for the particularly challenging setting of dense fog, making a significant step towards evaluation of semantic segmentation models in this setting.

In order to ensure a sound training and evaluation, we manually filter the unlabeled part of \emph{Foggy Zurich} and exclude from the resulting training sets those images which bear resemblance to any image in \emph{Foggy Zurich-test} with respect to the depicted scene.

\section{Experiments}
\label{sec:experiments}

\subsection{Fog Density Estimation with Synthetic Data}
\label{sec:exp:fog:density}

We conduct a user study on Amazon Mechanical Turk (AMT) to evaluate the ranking results of our fog density estimator. In order to guarantee high quality, we only employ AMT Masters in our study and verify the answers via a Known Answer Review Policy. Each human intelligence task (HIT) comprises five image pairs to be compared: three pairs are the true query pairs; the rest two pairs contain synthetic fog of different densities and are used for validation. The participants are shown two images at a time, side by side, and are simply asked to choose the one which is more foggy. The query pairs are sampled based on the ranking results of our method. In order to avoid confusing cases, \ie{}two images of similar fog densities, the two images of each pair need to be at least $20$ percentiles apart based on the ranking results.

We have collected answers for $12000$ pairs in $4000$ HITs. The HITs are only considered for evaluation only when both the validation questions are correctly answered. $87\%$ of all HITs are valid for evaluation. For these $10400$ annotations, we find that the agreement between our ranking method and human judgment is $89.3\%$. The high accuracy confirms that fog density estimation is a relatively easier task, and the solution to it can be exploited for solving high-level tasks of foggy scenes.

\subsection{Benefit of Adaptation with Our Synthetic Fog}
\label{sec:experiments:synthetic}

Our model of choice for experiments on semantic segmentation is the state-of-the-art RefineNet~\cite{refinenet}. We use the publicly available \emph{RefineNet-res101-Cityscapes} model, which has been trained on the clear-weather training set of Cityscapes. In all experiments of this section, we use a constant learning rate of $5\times{}10^{-5}$ and mini-batches of size 1. Moreover, we compile all versions of our synthetic foggy dataset by applying our fog simulation (which is denoted by ``Stereo-DBF'' in the following for short) on the same \emph{refined} set of Cityscapes images that was used in~\cite{SFSU_synthetic} to compile Foggy Cityscapes-refined. This set comprises 498 training and 52 validation images; we use the former for training. We considered dehazing as a preprocessing step as in~\cite{SFSU_synthetic} but did not observe a gain against \emph{no} dehazing and thus omit such comparisons from the following presentation.

Our first segmentation experiment shows that our semantic-aware fog simulation performs competitively compared to the fog simulation of~\cite{SFSU_synthetic} (denoted by ``Stereo-GF'') for generating synthetic data to adapt RefineNet to dense real fog. \emph{RefineNet-res101-Cityscapes} is fine-tuned on the version of Foggy Cityscapes-refined that corresponds to each simulation method for 8 epochs. We experiment with two synthetic fog densities. For evaluation, we use \emph{Foggy Zurich-test} as well as a subset of Foggy Driving~\cite{SFSU_synthetic} containing 21 images with dense fog, which we term Foggy Driving-dense, and report results in Tables~\ref{table:experiments:synthetic:foggy_zurich} and \ref{table:experiments:synthetic:foggy_driving_dense} respectively. Training on lighter synthetic fog helps to beat the baseline clear-weather model in all cases and yields consistently better results than denser synthetic fog, which verifies the first motivating assumption of CMAda at the end of Sec.~\ref{sec:CMAda}. In addition, Stereo-DBF beats Stereo-GF in most cases by a small margin and is consistently better at generating denser synthetic foggy data. On the other hand, Stereo-GF with light fog is slightly better for \emph{Foggy Zurich-test}. This motivates us to consistently use the model that has been trained with Stereo-GF in steps~\ref{itemfour} and \ref{itemfive} of CMAda for the experiments of Sec.~\ref{sec:experiments:curriculum}, assuming that its merit for dense real fog extends to lighter fog. However, Stereo-DBF is still fully relevant for step~\ref{itemsix} of CMAda based on its favorable comparison for denser synthetic fog.

\begin{table}[!tb]
  \centering
  \caption{Performance comparison on \emph{Foggy Zurich-test} of RefineNet and fine-tuned versions of it using Foggy Cityscapes-refined, rendered with different fog simulations and attenuation coefficients $\beta$}
  \label{table:experiments:synthetic:foggy_zurich}
  \setlength\tabcolsep{4pt}
  \begin{tabular}{lcc}
  \multicolumn{3}{c}{Mean IoU over \emph{all} classes (\%)}\\
  \toprule
  RefineNet~\cite{refinenet} & 32.0 &  \\
  \midrule
  Fog simulation & $\beta = 0.005$ & $\beta = 0.01$\\
  \midrule
  Stereo-GF~\cite{SFSU_synthetic} & \best{33.9} & 30.2\\
  Stereo-DBF & 33.4 & \best{31.2}\\
  \bottomrule
  \end{tabular}
  \hfil
  \begin{tabular}{lcc}
  \multicolumn{3}{c}{Mean IoU over \emph{frequent} classes (\%)}\\
  \toprule
  RefineNet~\cite{refinenet} & 48.8 &  \\
  \midrule
  Fog simulation & $\beta = 0.005$ & $\beta = 0.01$\\
  \midrule
  Stereo-GF~\cite{SFSU_synthetic} & \best{49.3} & 45.8\\
  Stereo-DBF & 49.0 & \best{46.6}\\
  \bottomrule
  \end{tabular}
\end{table}

\begin{table}[!tb]
  \centering
  \caption{Performance comparison on Foggy Driving-dense of RefineNet and fine-tuned versions of it using Foggy Cityscapes-refined, rendered with different fog simulations and attenuation coefficients $\beta$}
  \label{table:experiments:synthetic:foggy_driving_dense}
  \setlength\tabcolsep{4pt}
  \begin{tabular}{lcc}
  \multicolumn{3}{c}{Mean IoU over \emph{all} classes (\%)}\\
  \toprule
  RefineNet~\cite{refinenet} & 30.4 &  \\
  \midrule
  Fog simulation & $\beta = 0.005$ & $\beta = 0.01$\\
  \midrule
  Stereo-GF~\cite{SFSU_synthetic} & 32.5 & 32.4\\
  Stereo-DBF & \best{32.8} & \best{32.8}\\
  \bottomrule
  \end{tabular}
  \hfil
  \begin{tabular}{lcc}
  \multicolumn{3}{c}{Mean IoU over \emph{frequent} classes (\%)}\\
  \toprule
  RefineNet~\cite{refinenet} & 57.6 &  \\
  \midrule
  Fog simulation & $\beta = 0.005$ & $\beta = 0.01$\\
  \midrule
  Stereo-GF~\cite{SFSU_synthetic} & 60.4 & 58.7\\
  Stereo-DBF & \best{60.8} & \best{59.2}\\
  \bottomrule
  \end{tabular}
\end{table}

\subsection{Benefit of Curriculum Adaptation with Synthetic and Real Fog}
\label{sec:experiments:curriculum}

\begin{figure*}[!tb]
  \centering
  \begin{tabular}{ccccc}
    \includegraphics[width=0.24\textwidth]{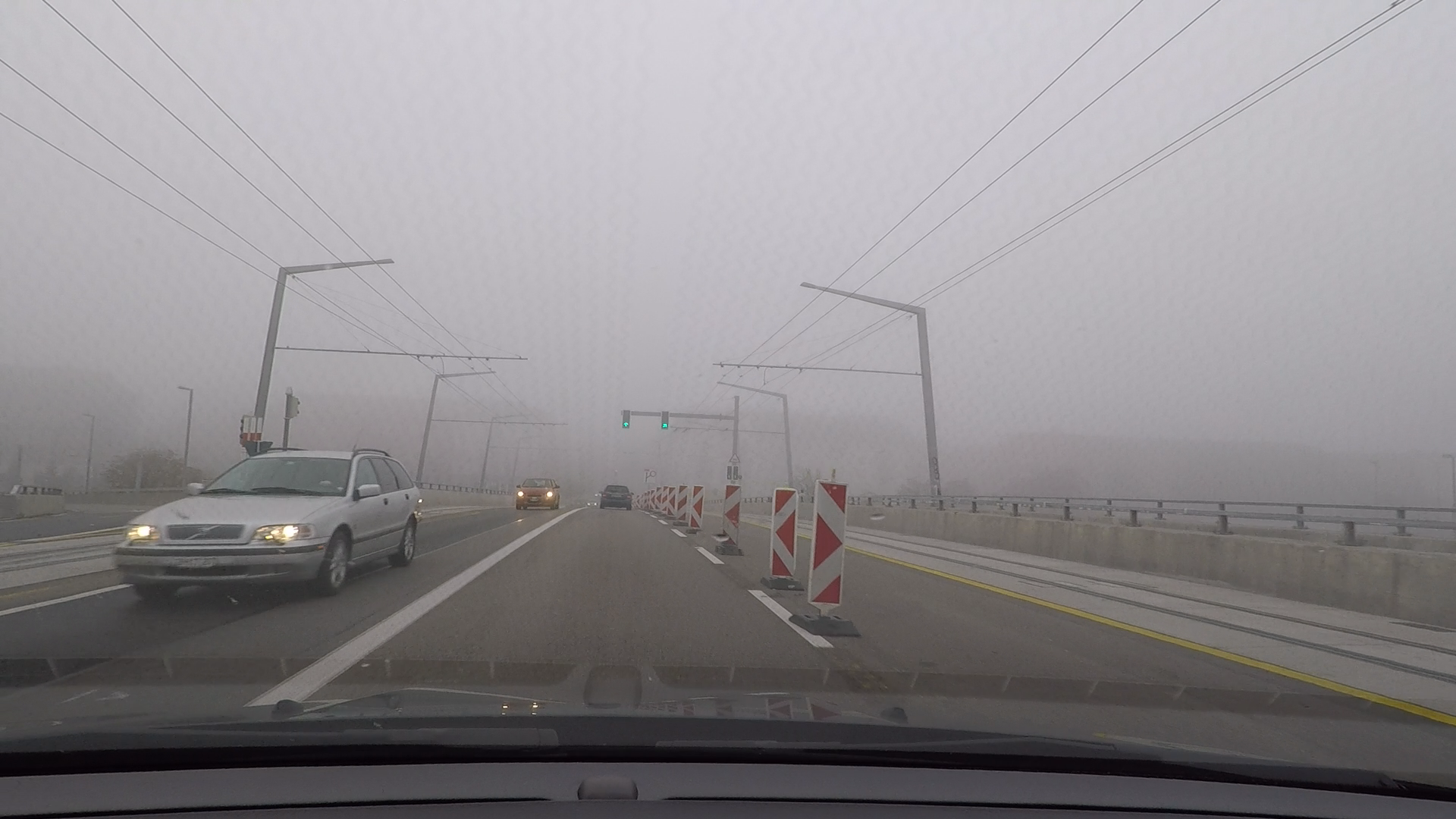}
    & \includegraphics[width=0.24\textwidth]{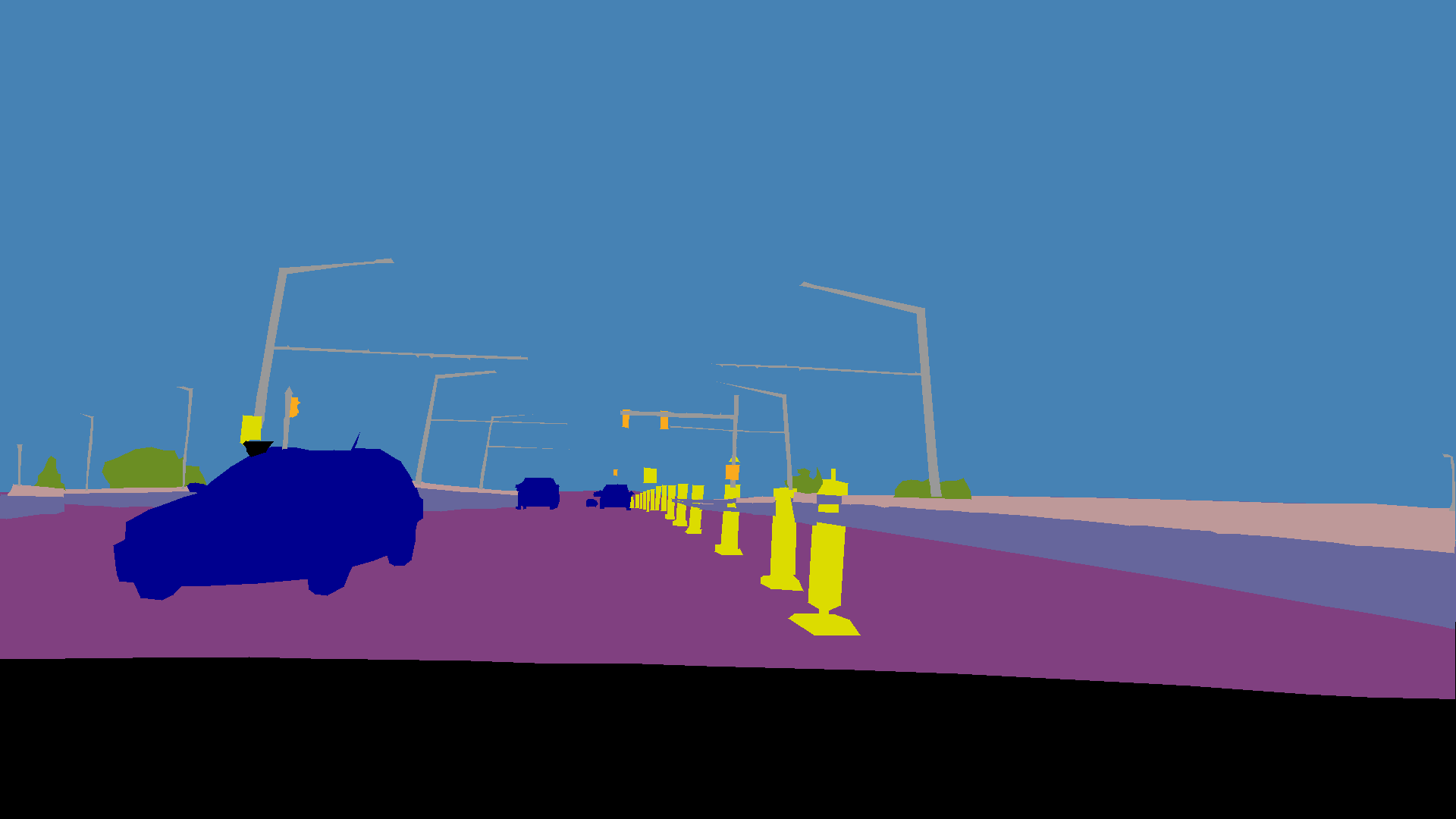}
    & \includegraphics[width=0.24\textwidth]{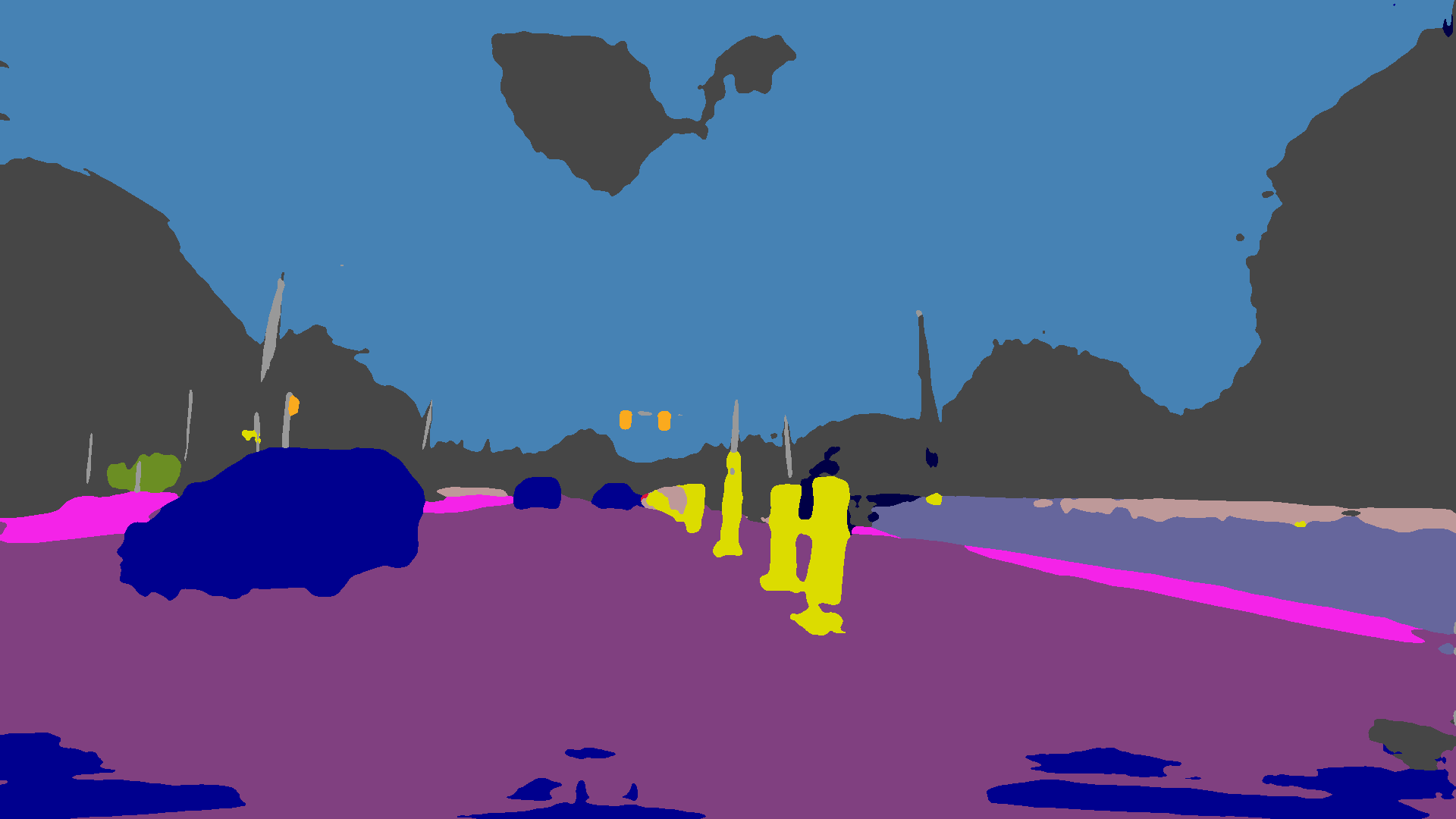}
    & \includegraphics[width=0.24\textwidth]{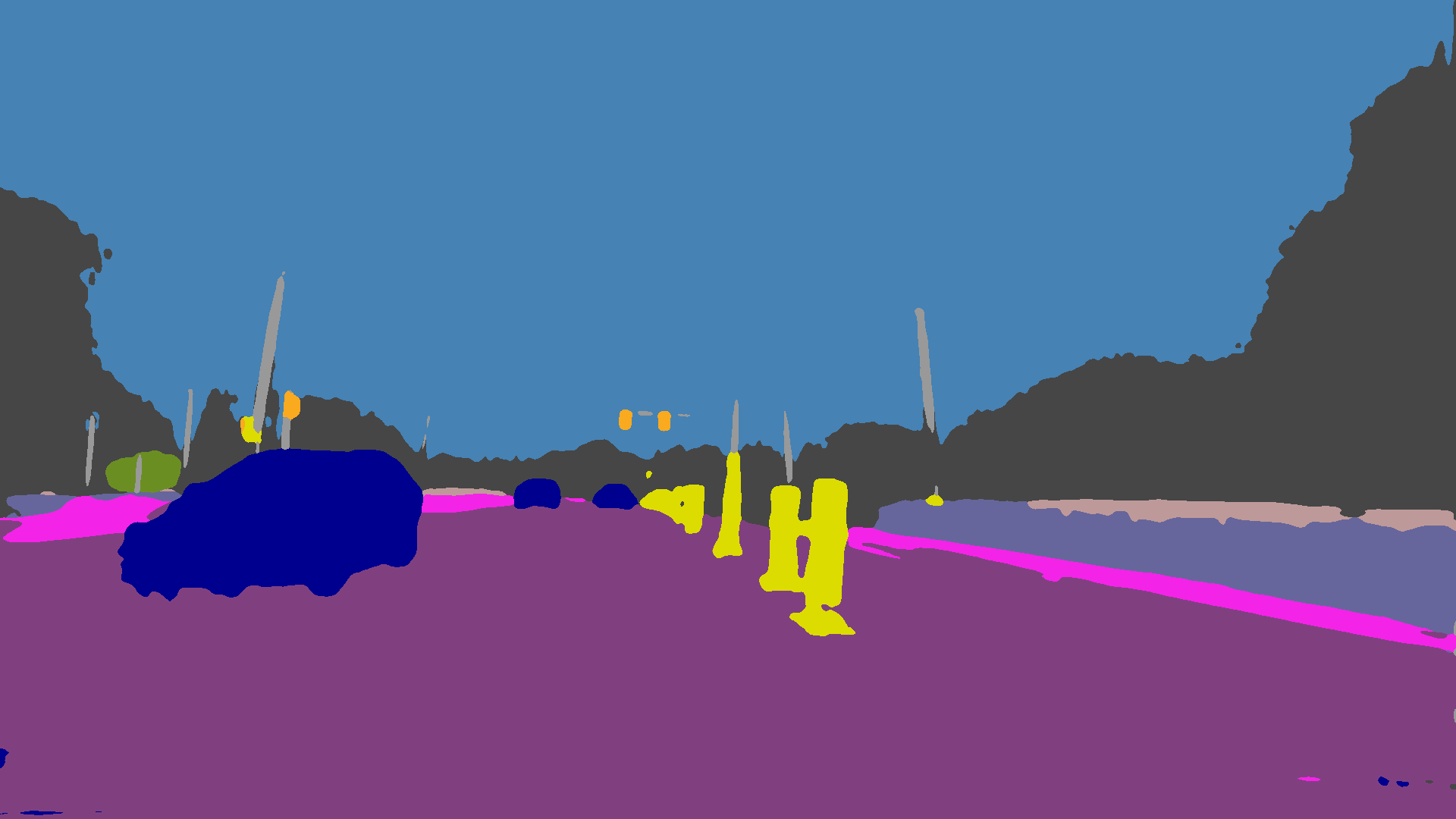}\\
    \includegraphics[width=0.24\textwidth]{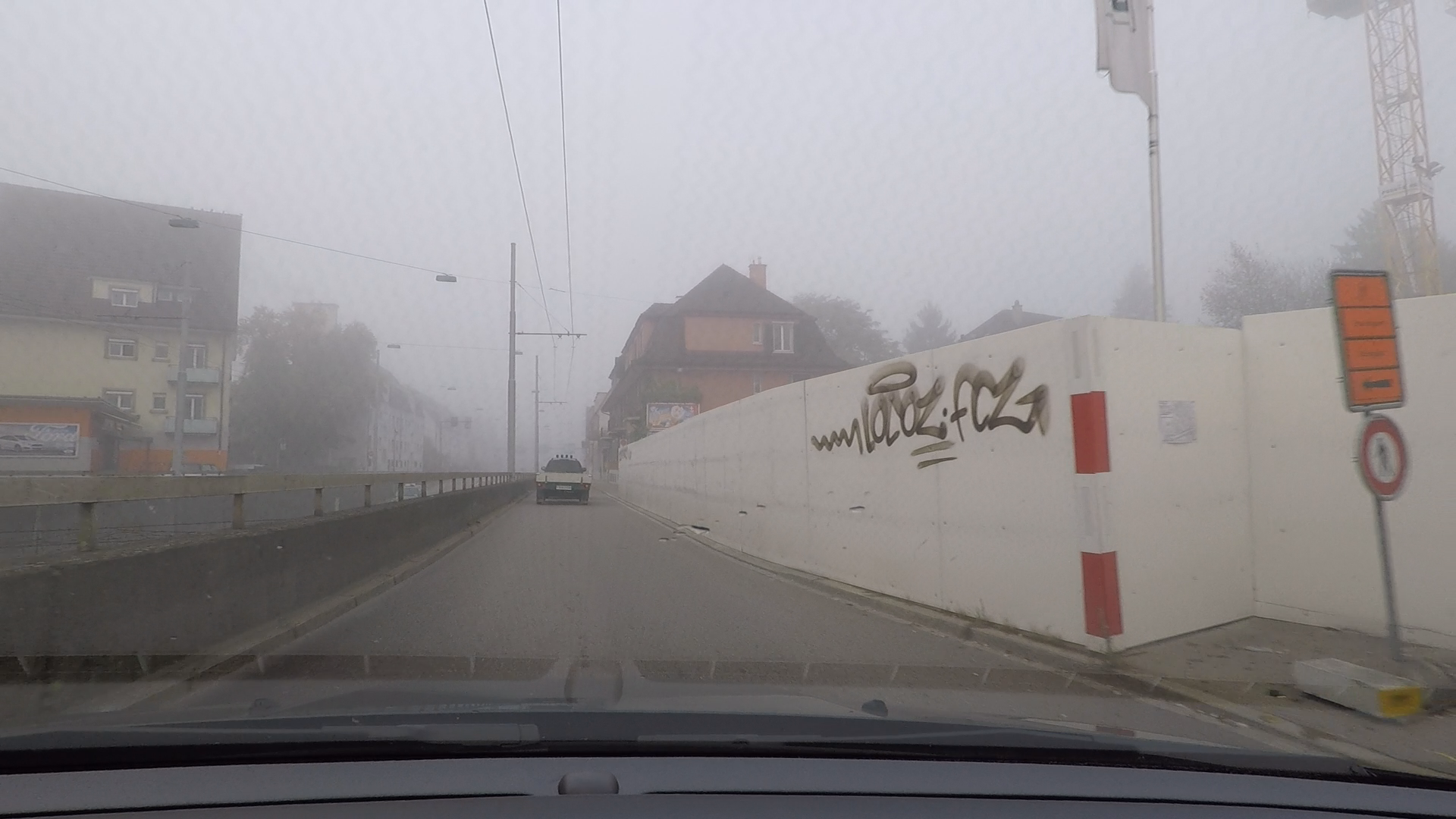}
    & \includegraphics[width=0.24\textwidth]{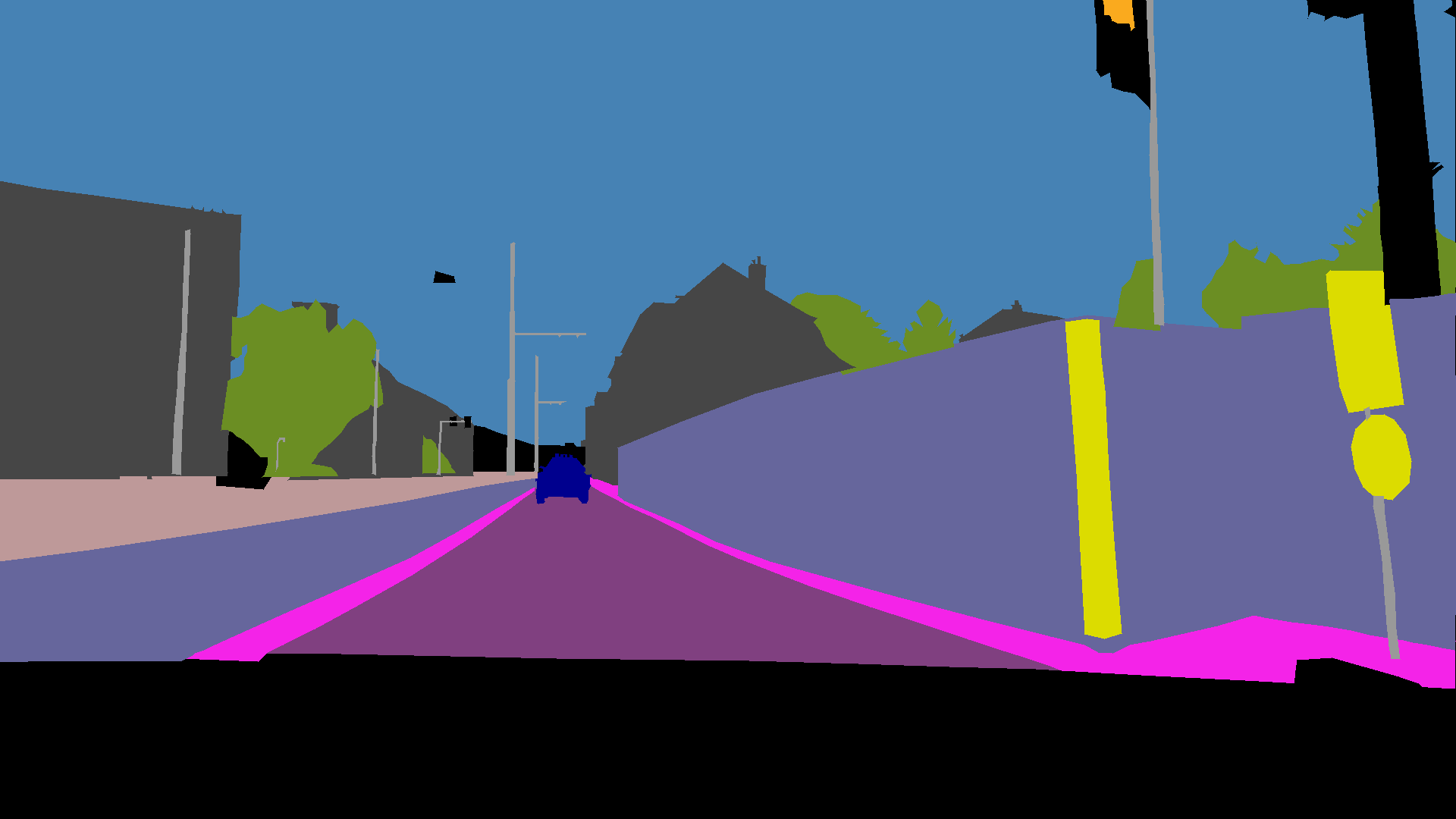}
    & \includegraphics[width=0.24\textwidth]{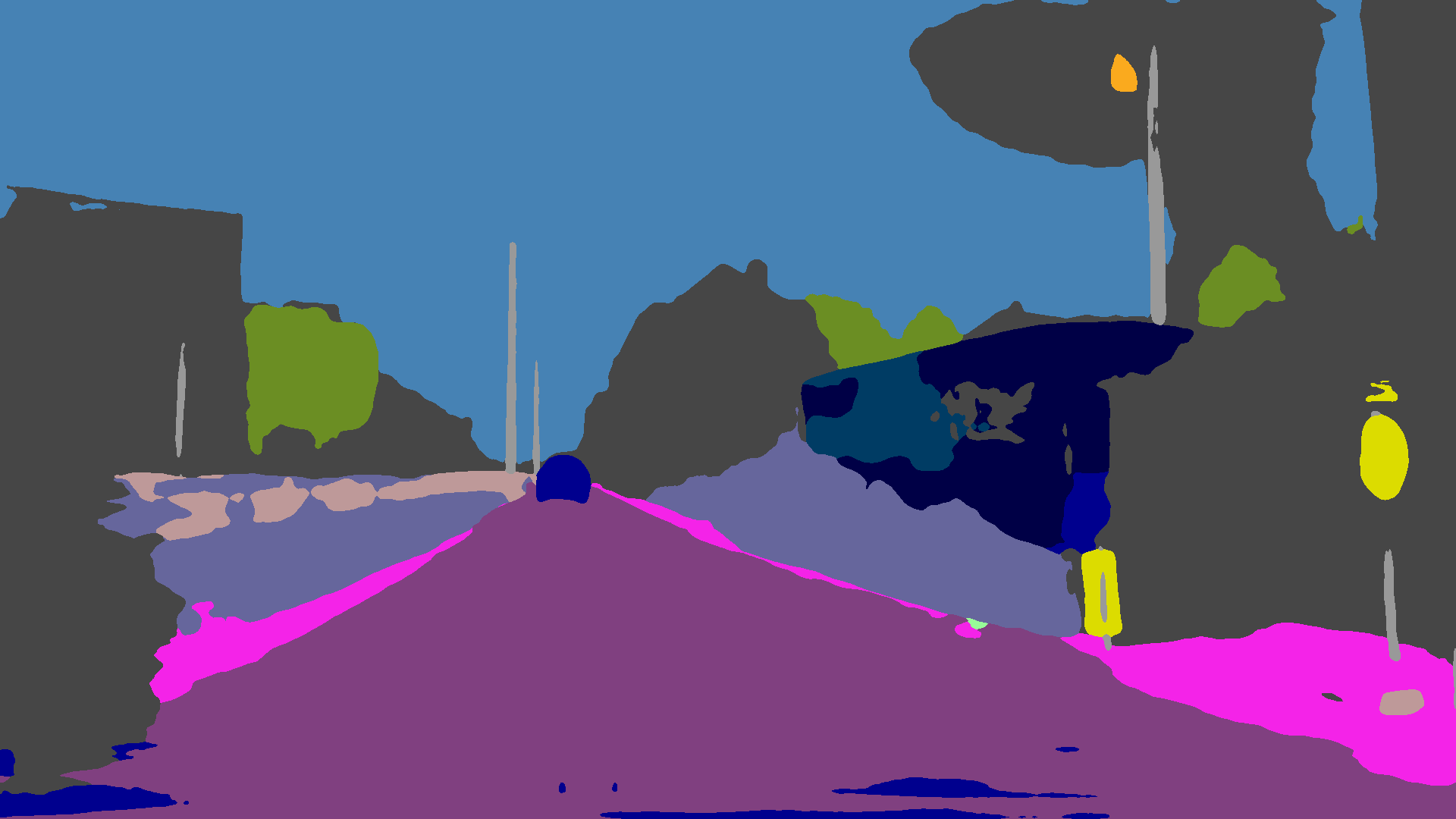}
    & \includegraphics[width=0.24\textwidth]{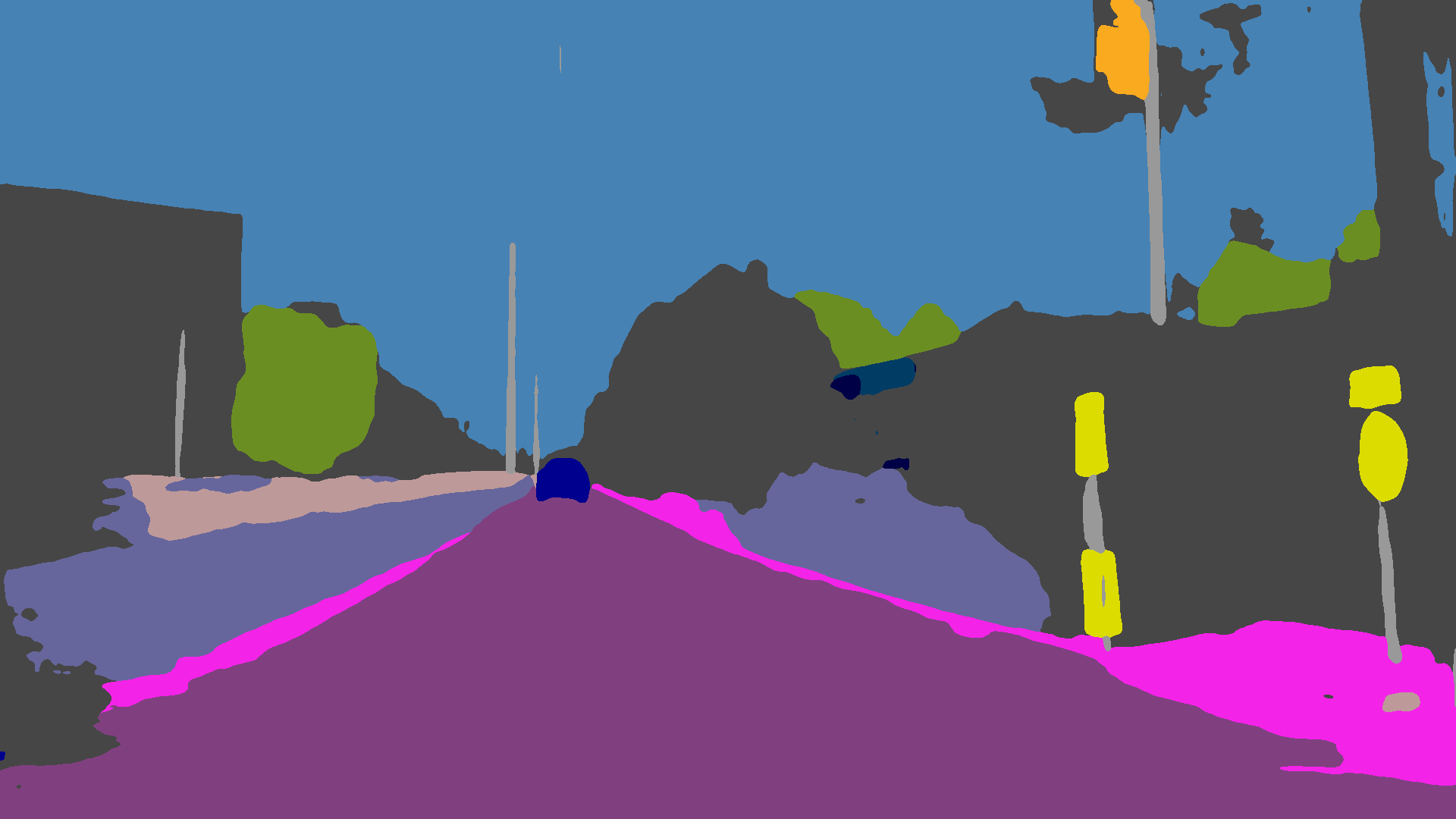}\\
    \includegraphics[width=0.24\textwidth]{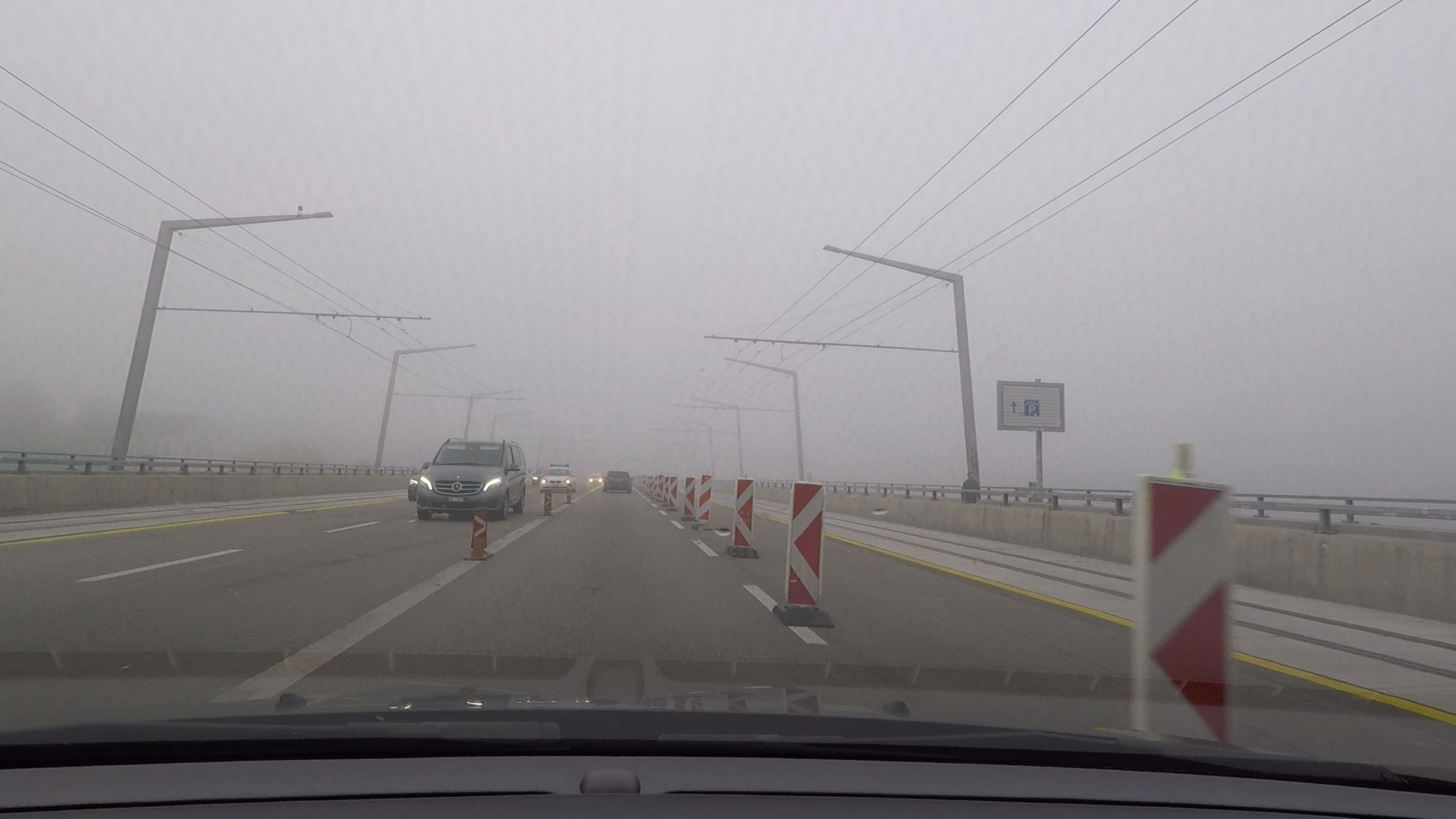}
    & \includegraphics[width=0.24\textwidth]{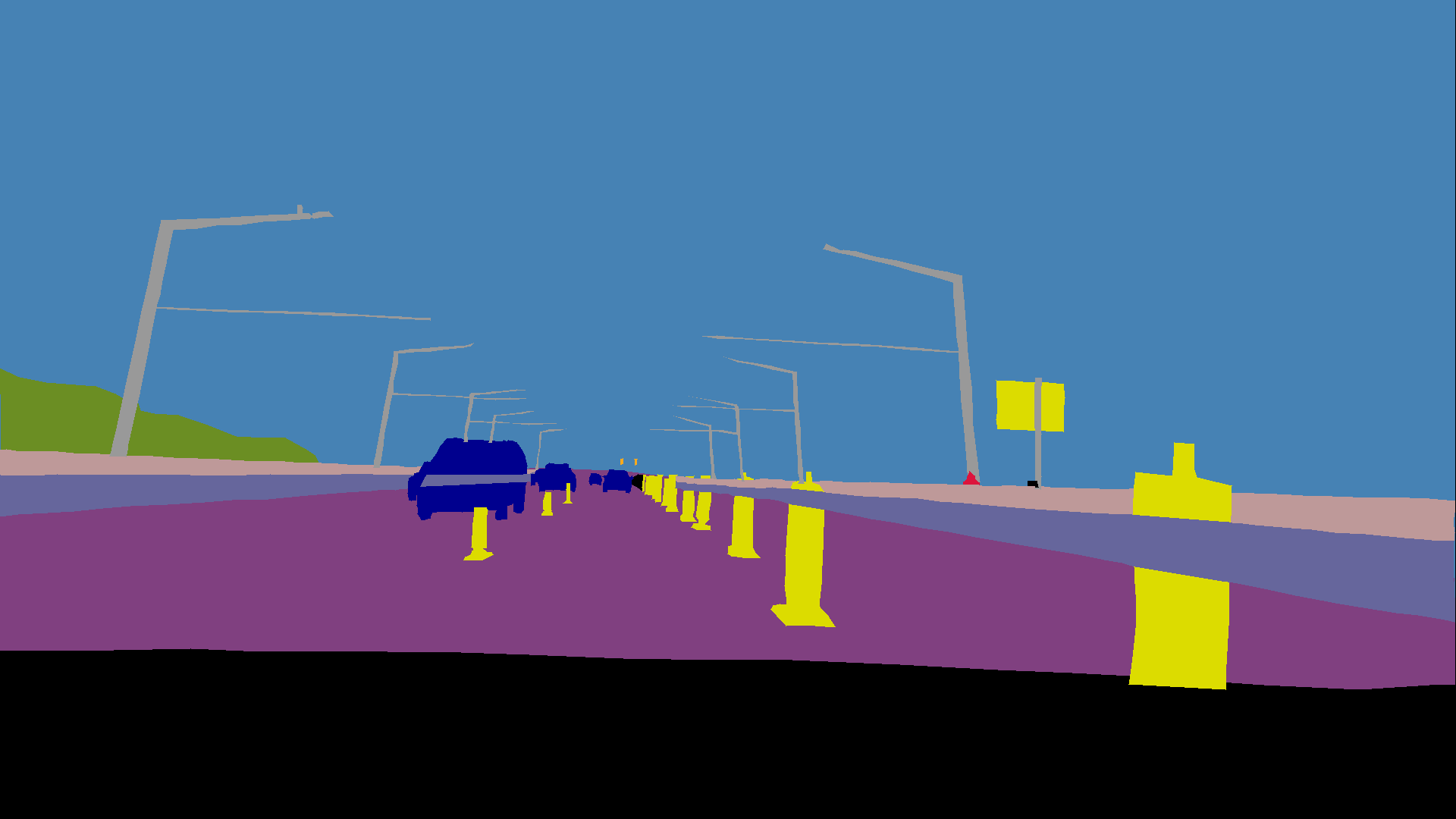}
    & \includegraphics[width=0.24\textwidth]{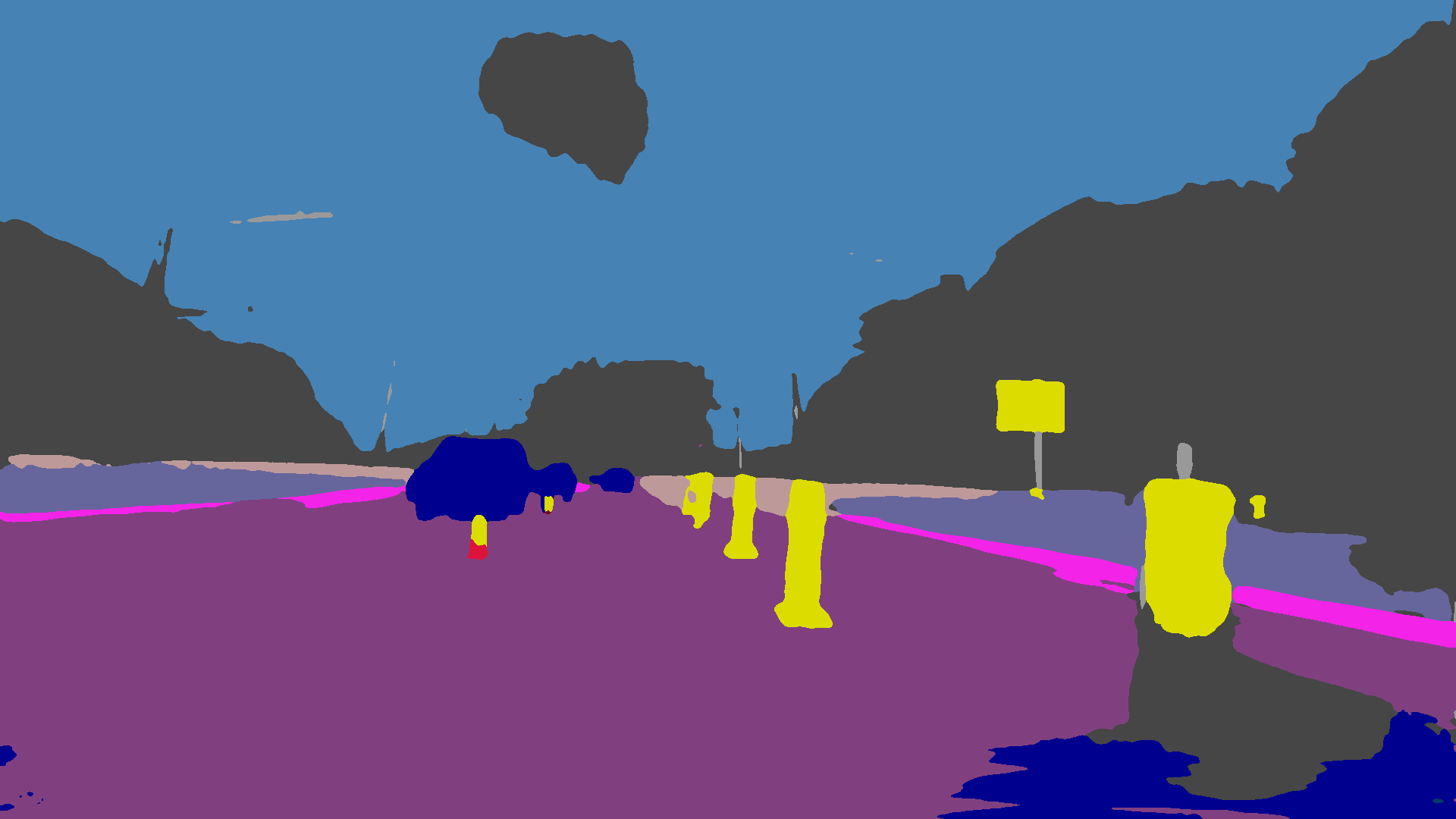}
    & \includegraphics[width=0.24\textwidth]{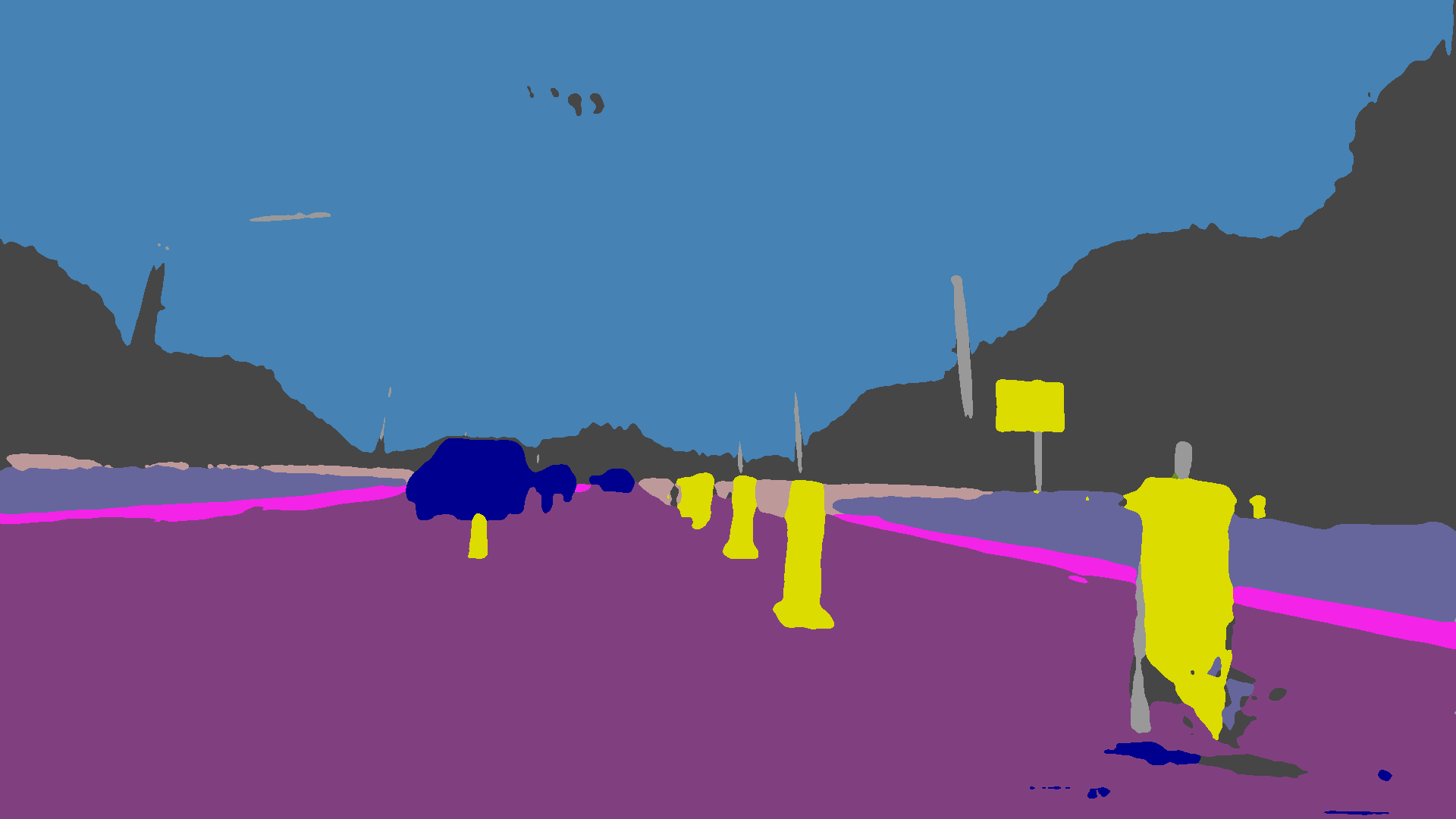}\\
    (a) foggy image & (b) ground truth & (c) RefineNet~\cite{refinenet} & (d) CMAda\\
  \end{tabular}
  \caption{Qualitative results for semantic segmentation on \emph{Foggy Zurich-test}. ``CMAda'' stands for RefineNet~\cite{refinenet} fine-tuned with our full CMAda pipeline on the union of Foggy Cityscapes-refined using our simulation and \emph{Foggy Zurich-light}}
  \label{fig:sem:seg}
\end{figure*}

Our second segmentation experiment showcases the effectiveness of our CMAda pipeline, using Stereo-DBF and Stereo-GF as alternatives for generating synthetic Foggy Cityscapes-refined in steps~\ref{itemfour} and \ref{itemsix} of the pipeline. \emph{Foggy Zurich} serves as the real foggy dataset in the pipeline. We use the results of our fog density estimation to select 1556 images with light fog and name this set \emph{Foggy Zurich-light}. The models which are obtained after the initial adaptation step that uses Foggy Cityscapes-refined with $\beta=0.005$ are further fine-tuned for 6k iterations on the union of Foggy Cityscapes-refined with $\beta=0.01$ and \emph{Foggy Zurich-light} setting $w=1/3$, where the latter set is noisily labeled by the aforementioned initially adapted models. Results for the two adaptation steps (denoted by ``CMAda-\ref{itemfour}'' and ``CMAda-\ref{itemseven}'') on \emph{Foggy Zurich-test} and \emph{Foggy Driving-dense} are reported in Tables~\ref{table:experiments:curriculum:foggy_zurich} and \ref{table:experiments:curriculum:foggy_driving_dense} respectively. The second adaptation step CMAda-\ref{itemseven}, which involves dense synthetic fog and light real fog, consistently improves upon the first step CMAda-\ref{itemfour}. Moreover, using our fog simulation to simulate dense synthetic fog for CMAda-\ref{itemseven} gives the best result on \emph{Foggy Zurich-test}, improving the clear-weather baseline by $5.9\%$ and $7.9\%$ in terms of mean IoU over all classes and frequent classes respectively. Fig.~\ref{fig:sem:seg} supports this result with visual comparisons. The real foggy images of \emph{Foggy Zurich-light} used in CMAda-\ref{itemseven} additionally provide a clear generalization benefit on Foggy Driving-dense, which involves different camera sensors than \emph{Foggy Zurich}.

\begin{table}[!tb]
  \centering
  \caption{Performance comparison on \emph{Foggy Zurich-test} of the two adaptation steps of CMAda using Foggy Cityscapes-refined and \emph{Foggy Zurich-light} for training}
  \label{table:experiments:curriculum:foggy_zurich}
  \setlength\tabcolsep{4pt}
  \begin{tabular}{lcc}
  \multicolumn{3}{c}{Mean IoU over \emph{all} classes (\%)}\\
  \toprule
  Fog simulation & CMAda-\ref{itemfour} & CMAda-\ref{itemseven}\\
  \midrule
  Stereo-GF~\cite{SFSU_synthetic} & 33.9 & 34.7\\
  Stereo-DBF & 33.4 & \best{37.9}\\
  \bottomrule
  \end{tabular}
  \hfil
  \begin{tabular}{lcc}
  \multicolumn{3}{c}{Mean IoU over \emph{frequent} classes (\%)}\\
  \toprule
  Fog simulation & CMAda-\ref{itemfour} & CMAda-\ref{itemseven}\\
  \midrule
  Stereo-GF~\cite{SFSU_synthetic} & 49.3 & 53.3\\
  Stereo-DBF & 49.0 & \best{56.7}\\
  \bottomrule
  \end{tabular}
\end{table}

\begin{table}[!tb]
  \centering
  \caption{Performance comparison on Foggy Driving-dense of the two adaptation steps of CMAda using Foggy Cityscapes-refined and \emph{Foggy Zurich-light} for training}
  \label{table:experiments:curriculum:foggy_driving_dense}
  \setlength\tabcolsep{4pt}
  \begin{tabular}{lcc}
  \multicolumn{3}{c}{Mean IoU over \emph{all} classes (\%)}\\
  \toprule
  Fog simulation & CMAda-\ref{itemfour} & CMAda-\ref{itemseven}\\
  \midrule
  Stereo-GF~\cite{SFSU_synthetic} & 32.5 & 34.1\\
  Stereo-DBF & 32.8 & \best{34.3}\\
  \bottomrule
  \end{tabular}
  \hfil
  \begin{tabular}{lcc}
  \multicolumn{3}{c}{Mean IoU over \emph{frequent} classes (\%)}\\
  \toprule
  Fog simulation & CMAda-\ref{itemfour} & CMAda-\ref{itemseven}\\
  \midrule
  Stereo-GF~\cite{SFSU_synthetic} & 60.4 & \best{61.6}\\
  Stereo-DBF & 60.8 & 61.5\\
  \bottomrule
  \end{tabular}
\end{table}

\section{Conclusion}
\label{sec:conclusion}

In this paper, we have shown the benefit of using partially synthetic as well as unlabeled real foggy data in a curriculum adaptation framework to progressively improve performance of state-of-the-art semantic segmentation models in dense real fog. To this end, we have proposed a novel fog simulation approach on real scenes, which leverages the semantic annotation of the scene as input to a novel dual-reference cross-bilateral filter, and applied it to the Cityscapes dataset. We have presented \emph{Foggy Zurich}, a large-scale dataset of real foggy scenes, including pixel-level semantic annotations for 16 scenes with dense fog. Through detailed evaluation, we have evidenced clearly that our curriculum adaptation method exploits both our synthetic and real data and significantly boosts performance on dense real fog without using any labeled real foggy image and that our fog simulation performs competitively to state-of-the-art counterparts.

\subsubsection{Acknowledgements.} This work is funded by Toyota Motor Europe via the research project TRACE-Z\"urich. 

\bibliographystyle{splncs04}
\bibliography{egbib}
\end{document}